\documentclass[preprint,review,12pt]{elsarticle}

\usepackage{amssymb}
\usepackage{amsmath}
\usepackage{graphicx}
\usepackage[retainorgcmds]{IEEEtrantools}
\usepackage[colorlinks]{hyperref}
\hypersetup{
  citecolor = {blue}
}
\usepackage{siunitx}
\usepackage{pdflscape}

\usepackage{enumitem}
\usepackage[normalem]{ulem}
\useunder{\uline}{\ul}{}
\usepackage{setspace} 
\usepackage[utf8]{inputenc}
\usepackage[linesnumbered, ruled, vlined]{algorithm2e}
\usepackage{rotating}
\usepackage{nicefrac}
\usepackage{xspace}
\usepackage{float}
\usepackage{caption}
\usepackage{subcaption}
\usepackage{rotating, lscape, longtable, tabu, booktabs, siunitx, multirow}
\usepackage[capitalise, noabbrev]{cleveref}
\usepackage{verbatim}
\usepackage{mathtools}
\usepackage{adjustbox} 
\usepackage{algpseudocode}
\usepackage{graphicx} 
\usepackage{booktabs} 
\usepackage{subcaption} 
\usepackage{tkz-graph}
\usepackage{geometry}
\usepackage{lineno} 
\usepackage{hyperref}
\usepackage{comment} 
\usetikzlibrary{arrows,arrows.meta,shapes,decorations.pathmorphing, decorations.markings, positioning}
\let\svtikzpicture\tikzpicture
\def\tikzpicture{\noindent\svtikzpicture}

\newcommand{\uset}[1]{\ifmmode\left\{\,#1\,\right\}\else\{\,#1\,\}\fi}
\newcommand{\ulst}[1]{\ifmmode\left[\,#1\,\right]\else[\,#1\,]\fi}
\newcommand{\upar}[1]{\ifmmode\left(\,#1\,\right)\else(\,#1\,)\fi}
\newcommand{\uioc}[1]{\ifmmode\left(\,#1\,\right]\else(\,#1\,]\fi}
\newcommand{\uico}[1]{\ifmmode\left[\,#1\,\right)\else[\,#1\,)\fi}

\journal{Advanced Engineering Informatics}

\begin{document}


\begin{frontmatter}

\title{
Scalable and reliable deep transfer learning for intelligent\\ fault detection via multi-scale neural processes\\ 
embedded with knowledge}

\author[unifal]{Zhongzhi Li}
\ead{zzli22@m.fudan.edu.cn}

\author[unifal]{Jingqi Tu}
\ead{jqtu23@m.fudan.edu.cn}

\author[unifal1]{Jiacheng Zhu}
\ead{zjc@csail.mit.edu}

\author[unifal]{Jianliang Ai}
\ead{aijl@fudan.edu.cn}

\author[unifal]{Yiqun Dong\corref{cor1}}
\ead{yiqundong@fudan.edu.cn}

\address[unifal]{Department of Aeronautics and Astronautics, Fudan University}
\address[unifal1]{Computer Science and Artificial Intelligence Laboratory, Massachusetts Institute of Technology}

\cortext[cor1]{Yiqun Dong is corresponding author}

\begin{abstract}
Deep transfer learning (DTL) is a fundamental method in the field of Intelligent Fault Detection (IFD). It aims to mitigate the degradation of method performance that arises from the discrepancies in data distribution between training set (source domain) and testing set (target domain). Considering the fact that fault data collection is challenging and certain faults are scarce, DTL-based methods face the limitation of available observable data, which reduces the detection performance of the methods in the target domain. Furthermore, DTL-based methods lack comprehensive uncertainty analysis that is essential for building reliable IFD systems. To address the aforementioned problems, this paper proposes a novel DTL-based method known as Neural Processes-based deep transfer learning with graph convolution network (GTNP). Feature-based transfer strategy of GTNP bridges the data distribution discrepancies of source domain and target domain in high-dimensional space. Both the joint modeling based on global and local latent variables and sparse sampling strategy reduce the demand of observable data in the target domain. The multi-scale uncertainty analysis is obtained by using the distribution characteristics of global and local latent variables. Global analysis of uncertainty enables GTNP to provide quantitative values that reflect the complexity of methods and the difficulty of tasks. Local analysis of uncertainty allows GTNP to model uncertainty (confidence of the fault detection result) at each sample affected by noise and bias. The validation of the proposed method is conducted across 3 IFD tasks, consistently showing the superior detection performance of GTNP compared to the other DTL-based methods. Beyond the application on IFD, GTNP holds broader implications and promises to enhance the development of reliable IFD systems in diverse scenarios, including autonomous driving, medical detection, etc.
\end{abstract}

\end{frontmatter}


\section{Introduction} \label{sec_into}
With the rapid development of deep learning (DL), intelligent fault detection (IFD) has claimed remarkable progresses in recent years \cite{khan2018review,huang2023intelligent,zhang2023realistic}. The success of DL-based IFD hinges on a core assumption: the training and testing data are drawn from the same distribution \cite{guo2018deep, zhao2021applications}. However, in real IFD applications, the data from the training and testing sets are collected under changing working conditions \cite{zhao2020intelligent} (e.g., rolling bearing data with different working loads) or on different machines \cite{li2023lightweight} (e.g., different aircraft), discrepancies in data distribution between the training and testing data occur frequently, resulting in the aforementioned assumption not being satisfied \cite{han2021hybrid}.

Deep transfer learning (DTL) is an effective method to address the aforementioned data distribution discrepancy problem. DTL aims to reuse the shared knowledge learned from one or more tasks to other related but different tasks \cite{weiss2016survey, iman2023review, li2020systematic}. In IFD, the source domain refers to one or multiple detection tasks that provide shared detection knowledge, while the target domain refers to other related tasks that the shared detection knowledge is applied to. In order to reuse the shared detection knowledge across different IFD tasks, researchers have developed many DTL-based methods, among which feature-based methods are the most widely adopted \cite{zhuang2020comprehensive,qian2022deep,li2023deep,zhang2022transfer,li2020intelligent,xiao2022deep}. This type of DTL-based methods aim to learn a mapping function to convert the data in source and target domains from the different data distributions to a shared latent data distribution to reduce the discrepancies.

Lu et al. proposed a feature-based DTL method for IFD, which introduces the Maximum Mean Discrepancy (MMD) term into the loss function to reduce the distribution discrepancies between different domains \cite{lu2016deep}. To further improve the detection performance, Wang et al. proposed Subdomain Adaptation Transfer Learning Network (SATLN), which combines subdomain adaptation with domain adaptation to reduce both marginal and conditional distribution biases simultaneously. Experiments carried out on 6 transfer tasks showed the SATLN achieved state-of-the-art detection results. Moreover, Qian et al. \cite{qian2018new} and An et al. \cite{an2020deep} utilized the KL Divergence and CORrelation ALignment (CORAL) to minimize the distribution discrepancies between source and target domains, respectively. Li et al. \cite{xiong2021multi} employed Central Moment Discrepancy (CMD) to reduce the discrepancies between different domains, which yields remarkable detection accuracy. Nevertheless, these methods assume that there is sufficient data in the target domain. Due to the complex of collecting processes (e.g., rolling bearing data with changing working loads), strict legal rules around data privacy and compliance (e.g., aircraft flight data), and the unpredictable occurrence of emerging faults, data in the target domain are often limited \cite{wang2019domain}, which jeoperdizes the performance of DTL-based methods.

Furthermore, a reliable IFD system should not only focus on detection accuracy, but also on the confidence level of the outputs. Since traditional DTL-based methods typically rely on observable data distributions in the target domain only, the discrepancy between the observable data distribution and the inherent distribution in the target domain emerges in complex environments \cite{han2022out, shim2021fault}. The distribution discrepancy in the target domain arises from factors such as noise during data collection (e.g., in aircraft fault data) and equipment degradation (e.g., in rolling bearing fault data), which can raise the risk of failing to detect these safety-critical faults \cite{shamsi2021uncertainty}. However, current DTL-based IFD methods failed to analyze the uncertainty of detection outputs comprehensively \cite{li2022perspective}.

In recent years, researchers have made efforts in estimating uncertainty for a single domain in the IFD field \cite{cai2017bayesian}. One potential way is Bayesian Neural Network (BNN) \cite{maged2022uncertainty,yongli2006bayesian,sun2020fault}, which assumes a prior distribution over the weights of neural network that can represent the uncertainty in the posterior distribution after model training. However, inference of neural network weights for IFD tasks can be difficult due to the high dimensionality and posterior complexity \cite{louizos2017multiplicative}. Gaussian processes (GPs) \cite{liang2021probabilistic,du2018fault} are further adopted, which offer a direct approach to assume distributions over neural networks, without the necessity of utilizing prior distributions on neural network weights. Therefore, for regular GPs, posterior inference is considerably simpler. Despite these advantages, GPs exist 2 primary limitations when used for IFD tasks: 1) the fixed kernel function lacks the flexibility to handle high-dimensional data, and 2) the training and testing of GPs are quite costly, typically scaling cubically with the size of the dataset \cite{haftka2016parallel}.

The Neural Processes (NP) proposed by DeepMind is an advanced method to overcome the limitations of GPs, which combines the advantages of neural networks and GPs \cite{garnelo2018neural}. On the one hand, NP has powerful representation ability due to utilizing neural network, which learns an implicit kernel function (a neural network with trainable parameters), thus avoiding the limitation of using the fixed kernel function in GPs; on the other hand, the NP utilizes the uncertainty modeling method of GPs, which can provide uncertainty estimation of the fault detection results \cite{nguyen2022transformer}, reducing the occurrence of the false positives or false negatives. However, current NP-based methods (NPs) still lack a comprehensive analysis of uncertainty. Several of these methods rely on global latent variables \cite{garnelo2018conditional,wang2021global}, which reflect the characteristics or trend of the entire dataset. Due to the constraints imposed by increasing computational complexity, global latent variables demonstrate limited applicability for large-scale data. Other methods depend on local latent variables \cite{louizos2019functional}, which are affected by specific data or subsets, resulting in insufficient capture of the data distribution of the entire dataset. In Attentive Neural Processes (ANPs) \cite{kim2019attentive}, the performance of ANPs (joint modeling with different attention mechanisms) on various tasks is enhanced by the employment of \textit{self-attention}, which is applied to compute representations of each ($x, y$) pair, and \textit{cross-attention}, which focuses on context representations. However, in terms of uncertainty, ANPs only provide the uncertainty associated with the final results, without providing estimates of uncertainties at model level and task level.

Based on the above analysis, we proposed NP-based deep transfer learning with the graph convolution network (GTNP) method, as shown in Figure \ref{fig_A2}. GTNP is capable of performing fault detection in complex environments and also provide more comprehensive results of uncertainty analysis. To address the challenge of limitations of observable data in the target domain, GTNP has 3 improvements: 1) constructing a graph convolutional network (GCN) embedding the detection knowledge of source domain; 2) joint modeling strategy using global and local latent variables; 3) conducting sparse sampling in the target domain. To address the uncertainty estimates, GTNP constructs a multi-scale uncertainty analysis approach at the model level and sample level for the IFD tasks. The main highlights of this paper are summarized as follows:

\begin{figure}[!t]
\centering
\includegraphics[width=6in]{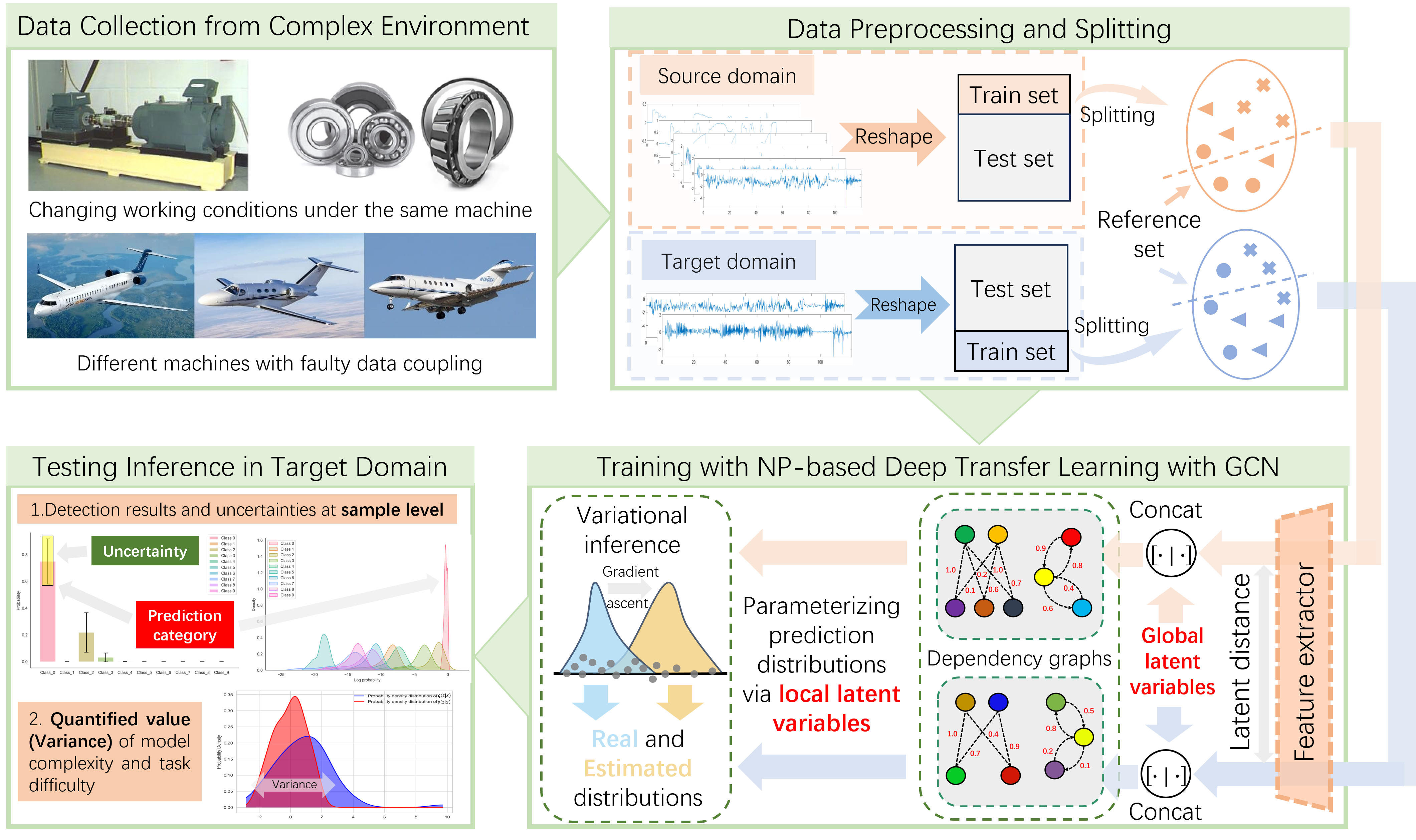}
\caption{Schematic diagram of applying the proposed GTNP to IFD tasks in complex environment.}
\label{fig_A2}
\end{figure}

(1) \textbf{Novel Deep Transfer Learning Framework for IFD:} GTNP method, which combines the benefits of DTL (high accuracy and scalability) and NPs (reliability), has been proposed. GTNP achieves the fault detection tasks with limited data in the target domain while realizing a multi-scale uncertainty analysis on both model and sample levels, which embodies both the scalability and reliability for IFD tasks.

(2) \textbf{Efficient Modeling Strategy with Limited Data in Target Domain:} Firstly, GCN is used to represent the detection knowledge extracted from the source domain to improve the detection performance of GTNP in the target domain \cite{lin2023coronary}. Furthermore, the dependence graph constructed by the GCN enriches the context information for the inference process of GTNP , which further improving the detection accuracy. Secondly, a global and local joint modeling strategy has been proposed, in which global latent variables impart insight into the entire data distribution and facilitate the transfer of detection knowledge; while local latent variables promote the model's adaptation to the specific data distribution of the target domain, especially when there are large discrepancies between the source and target domains. Lastly, a sparse sampling strategy named reference set is introduced into GTNP to reduce the demand of the observable data in the target domain. It is similar to the ‘‘inducing inputs’’ in Sparse GPs \cite{titsias2009variational} that use a limited amount of data to estimate the mean and variance of the entire data distribution of the target domain.

(3) \textbf{Multi-scale Analysis of Uncertainty for IFD:} The method of joint modeling strategy demonstrates the capability of the proposed GTNP model to conduct a multi-scale analysis of uncertainty. The uncertainty analysis includes global uncertainty, which operates at the model level and quantifies the complexity of the models and the difficulty of the tasks. Besides, local uncertainty, operating at the sample level, provides uncertainty estimation for a single sample.

(4) \textbf{Validation of The Proposed Method on Multiple Tasks:} To verify the efficiency of GTNP, this paper designs 3 transfer tasks, namely rolling bearing fault detection under different working loads, sensor fault detection between different aircraft, and detection of emerging fault that only exist in the target domain, which correspond to the typical tasks in practical IFD applications. Comparison experiments with other advanced DTL-based methods show that the GTNP method proposed in this paper achieves the highest detection accuracy and can provide comprehensive uncertainty analysis at the model level and sample level.

\section{Problem statement} \label{sec:problem_statement}
The IFD for cross domains is generally defined as a task of applying the detection knowledge from source domain to target domain. Specifically, the source domain is defined as $D_{s}=\left\{X_{s}, P_{\mathrm{s}}\left(X_{\mathrm{s}}\right)\right\}$. The sample space in source domain $D_{s}$ is $X_{s}=\left\{x_{s}^{i}, y_{s}^{i}\right\}_{i=1}^{n_{s}}$, where $x_{s}^{i}$ denotes the $i^{th}$ input data, $y_{s}^{i}$ denotes the $i^{th}$ label and $n_{s}$ denotes the total number of samples. $D_{s}$ comprises 2 integral components: the sample space $X_{s}$ and its probability distribution $P_{\mathrm{s}}\left(X_{\mathrm{s}}\right)$. Similarly, the target domain is regarded as $D_{t}=\left\{X_{t}, P_{\mathrm{t}}\left(X_{\mathrm{t}}\right)\right\}$. There are $n_t$ samples in $X_{t}=\left\{x_{t}^{i}, y_{t}^{i}\right\}_{i=1}^{n_{t}}$ from the target domain with the distribution of $P_{\mathrm{t}}\left(X_{\mathrm{t}}\right)$. And it should be noted that $P_{\mathrm{t}}\left(X_{\mathrm{t}}\right) \neq P_{\mathrm{s}}\left(X_{\mathrm{s}}\right)$ due to the samples in these two domains are collected from different conditions or machines. Given the label space $\Psi=\{1,2, \ldots, \psi\}$ that contains $\psi$ classes, the labels of source domain and target domain are $y_{s}^{{i}} \in \Psi^{\mathrm{s}} \subseteq \Psi$ and $y_{t}^{{i}} \in \Psi^{\mathrm{t}} \subseteq \Psi$, respectively. The DTL for IFD task is defined as $f^{S \rightarrow T}(\cdot):= X_{s} \rightarrow {X}_{t}$, which aims to find a mapping function $f^{S \rightarrow T}(\cdot)$ that transfers the detection knowledge from the source domain to the target domain, as shown in Figure \ref{fig_A1}. In this paper, in order to evaluate the effectiveness of the proposed method, we conduct 2 application scenarios of IFD on 3 tasks:

\begin{figure}[!t]
\centering
\includegraphics[width=6in]{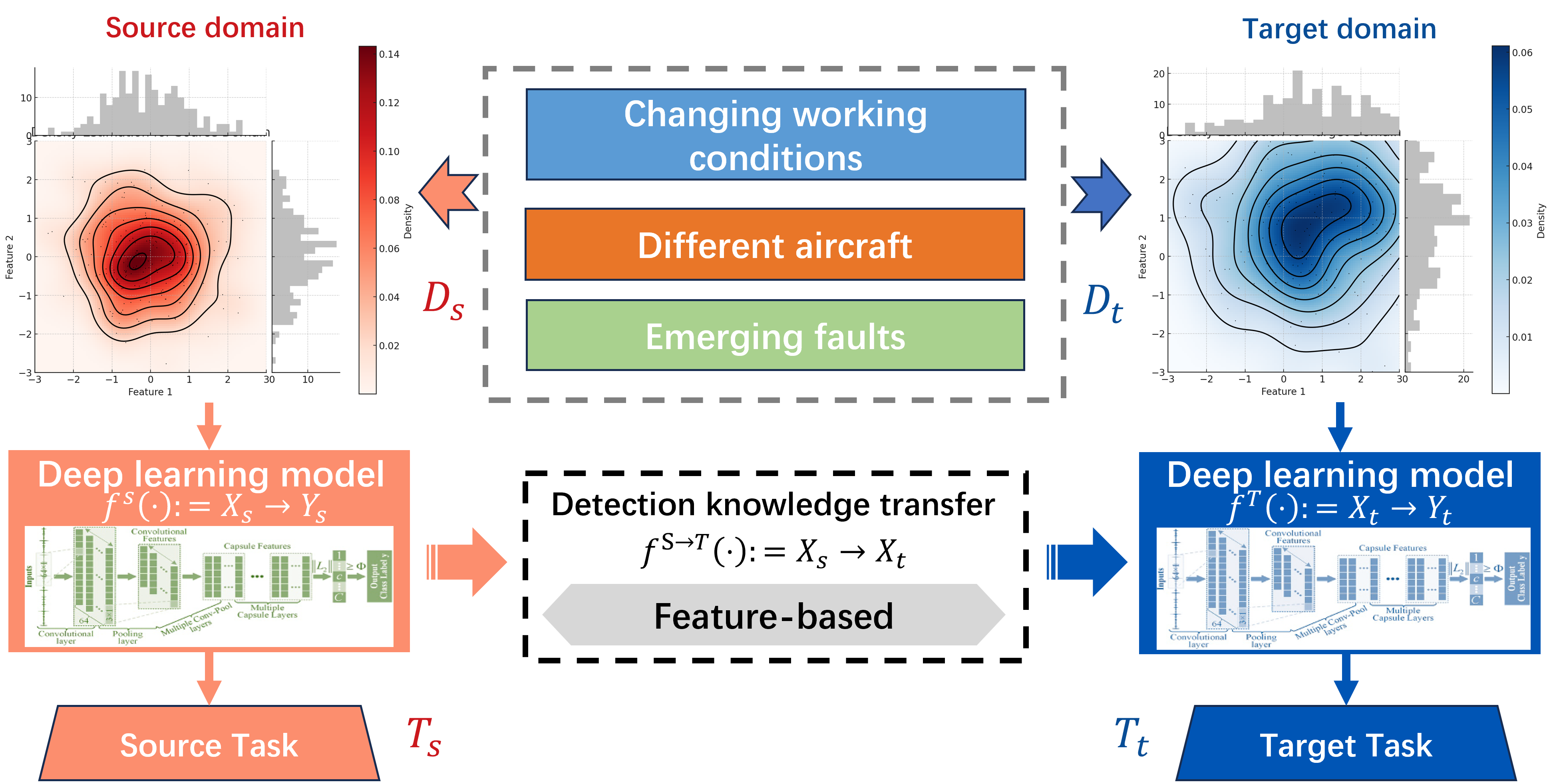}
\caption{The transfer principle of DTL for IFD tasks in this paper.}
\label{fig_A1}
\end{figure}

\begin{figure*}[!h] 
    \centering
    \begin{subfigure}[b]{0.48\textwidth}
        \includegraphics[width=\textwidth]{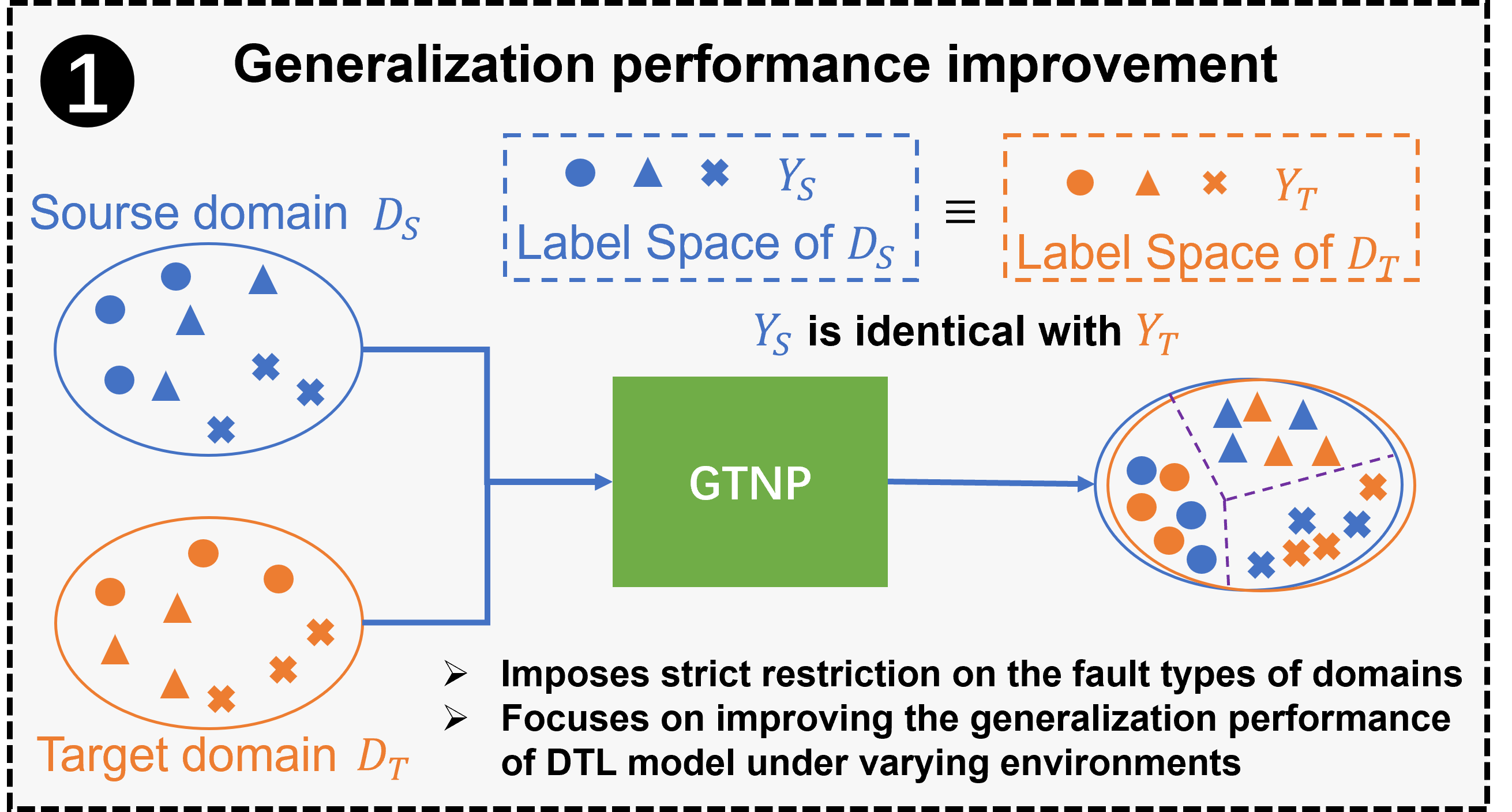}
        \caption{}
        \label{E_1}
    \end{subfigure}
    \begin{subfigure}[b]{0.48\textwidth}
        \includegraphics[width=\textwidth]{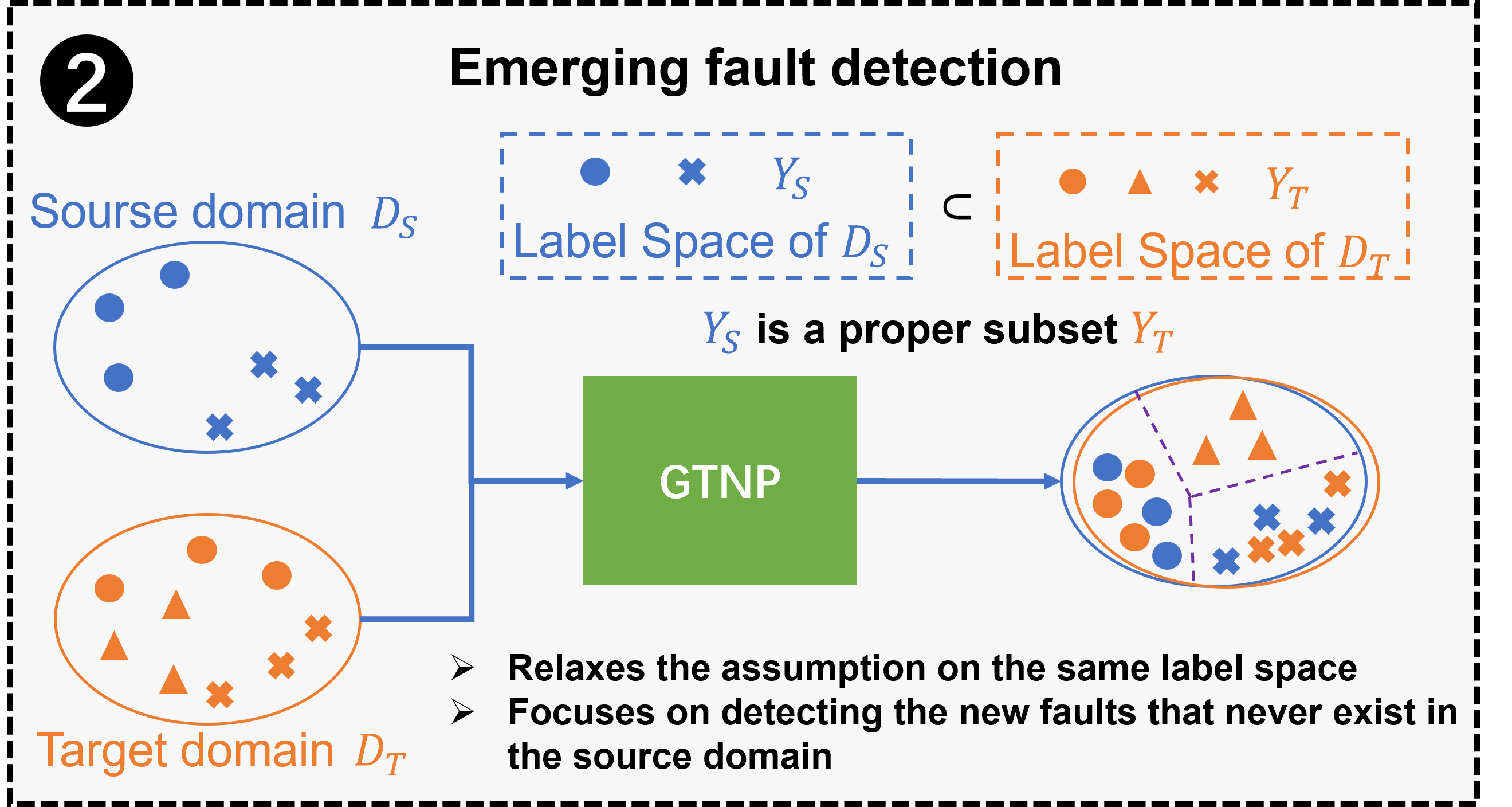}
        \caption{}
        \label{E_2}
    \end{subfigure}
    \caption{Illustration of the 2 IFD scenarios for GTNP. (a) Generalization Performance Improvement, (b) emerging Fault Detection.}
    \label{fig_1}
\end{figure*}

\textbf{Generalization performance improvement:} in this scenario, the label space of target domain is identical with the label space of source domain ($\Psi^{\mathrm{s}} \equiv \Psi^{\mathrm{t}}$). The purpose of the experiments are mainly focuses on improving the generalization performance of DTL under varying environments, as shown in Figure \ref{E_1}. Two experiments are conducted in this scenario: 1) the complexity of the operating parameters of rolling bearing machines exacerbates the challenge of detection knowledge transfer between domains. We verify the performance of GTNP under cross-working conditions in the bearing fault detection task. 2) In contrast to the changing working conditions, cross-machines IFD tasks involve data collected from related yet distinct machines, influenced by intricate factors like mechanical structures and materials. We test the ability of GTNP to transfer detection knowledge between different aircraft in the context of sensor fault detection. In particular, we focus on evaluating the model's performance in transferring detection knowledge from simulated aircraft (source domain) to real aircraft (target domain).

\textbf{Emerging fault detection:} in this scenario, the label space of the target domain is a superset of the label space of the source domain ($\Psi^{\mathrm{s}} \subseteq \Psi^{\mathrm{t}}$), which mainly focuses on detecting the new faults that never exist in the source domain, as shown in Figure \ref{E_2}. In real-world tasks, the emergence of short-term fault is a common occurrence since the machine operates in intricate and unpredictable environments. Consequently, it is necessary for GTNP to promptly detect the emerging fault in practical IFD applications. Mathematically, the detection process of GTNP can be described as:
\begin{equation}
    \left[y_{s}^{i}, y_{t}^{j}\right]=\operatorname{GTNP}\left(x_{s}^{i}, x_{t}^{j}\right), i \in\left[1, n_{s}\right], j \in\left[1, n_{t}\right]
    \label{eq:2.1}
\end{equation}

More details about the proposed GTNP can be found in Section \ref{sec:method}.

\section{Methodology} \label{sec:method}
\subsection{The proposed GTNP framework}
\begin{figure}[!t]
\centering
\includegraphics[width=6in]{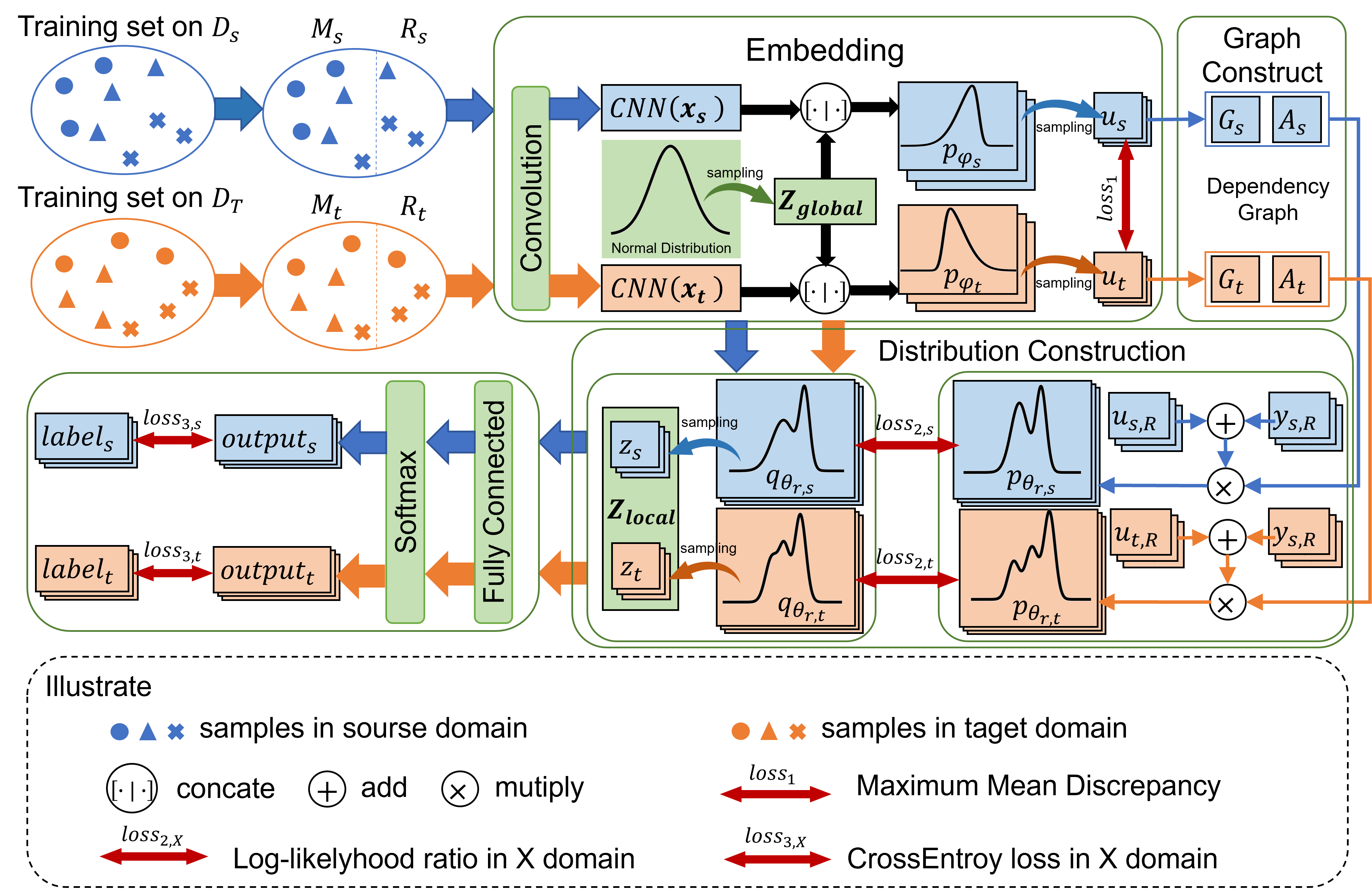}
\caption{The framework of GTNP proposed in this paper.}
\label{fig_12}
\end{figure}

The overall framework of the proposed GTNP is shown in Figure \ref{fig_12}, including an Embedding module, a Graph Construct module, and a Distribution Construction module. Unlike traditional DTL methods using the feature vector of the input data directly, GTNP uses the feature vector of the input data to establish its unique distribution which can provide uncertainty information and solve the IFD tasks from the perspective of distribution. To address the problem of sparse observable data in the target domain, we first split the sample in a domain into R set (reference set) and M set (training set except reference set) inspired by ‘‘inducing inputs’’ in Sparse GPs. In the subsequent formula, we use the subscripts $s$ and $t$ to represent the source domain and target domain, respectively.

\begin{equation}
	{u}_{{s}}, {u}_{{t}}=\text { Embedding }\left({x}_{{s}}, {x}_{{t}}\right)
 \label{Embedding}
\end{equation}

Firstly, in Embedding module, GTNP embeds the data of the input sample into a latent space, which can capture latent structure and patterns in the data. As shown in equation \ref{Embedding}, ${x_s}$, ${x_t}$ represent the data of the input sample from 2 domains, ${u_s}$, ${u_t}$ are the latent embedding representations of input data in the latent space.  ${x_s}$, ${x_t}$ are first input into a shared weight feature extractor. Subsequently, the global latent variable ${z_{global}}$ sampled from the normal distribution is concatenated with the outputs of the feature extractor. We use the outputs of the previous 2 operations on each sample to construct a sample-specific distribution.

The next step is to construct the dependency graphs encoding the correlation structures in Graph Construction module. Unlike GPs utilizing a kernel function to encode the correlation structures in a covariance matrix, GTNP adopts a distinct approach that constructs the dependency graphs in R set and between R set and M set, which are respectively defined as G and A, using the latent embedding ${u_s}$ and ${u_t}$. G is constructed by GCN to represent the dependencies between observable samples in the R set, while A is constructed by a bipartite graph between R set and M set to represent the dependencies between observable and expanded samples in domains, which enhances the extrapolation ability of the GTNP model.
\begin{equation}
	G_{s}, A_{s}, G_{t}, A_{t}=\text { GraphConstruction }\left(\left({u}_{s, {R}}, {u}_{s, {M}}\right),\left({u}_{t, {R}}, {u}_{t, {M}}\right)\right)
 \label{GraphConstruction}
\end{equation}

Having obtained the dependency graphs G and A, in Distribution Construction module, we first construct a distribution based on G, A, and samples in the R set, which is defined as the real distribution of the samples in the domain. GTNP model also constructs a distribution only based on the samples in the R set, which is defined as the estimated distribution of the samples in the domain.

\begin{equation}
\begin{array}{c}
{p}_{\theta_{r, s}}\left({Z}_{{s}} \mid G_{s}, A_{s}, {U}_{{s}, {R}}, {Y}_{{s}, {R}}\right), {p}_{\theta_{r, t}}\left({Z}_{{t}} \mid G_{t}, A_{t}, {U}_{{t}, {R}}, {Y}_{{t}, {R}}\right) \\
=\text { Distribution }_{r}
\left({z}_{{s}}, {z}_{{t}} \mid\left(G_{s}, A_{s}, {u}_{{s}, {R}}, {y}_{{s}, {R}}\right),\left(G_{t}, A_{t}, {u}_{{t}, {R}}, {y}_{{t}, {R}}\right)\right)
\end{array}
 \label{Distributione}
\end{equation}

\begin{equation}
{q}_{\theta_{e, s}}\left({Z}_{{s}} \mid {U}_{{s}}\right), {q}_{\theta_{e, t}}\left({Z}_{{t}} \mid {U}_{{t}}\right)=\text { Distribution }_{e}\left({z}_{{s}}, {z}_{{t}} \mid {u}_{{s}}, {u}_{{t}}\right)
 \label{Distributitwo}
\end{equation}

In equation \ref{Distributione} and \ref{Distributitwo}, ${p}$, ${q}$ represent the real distribution and the estimated distribution, respectively. $\theta_{r}$, $\theta_{e}$ represent the parameters of the 2 distributions. $z$ represents the local latent variable corresponding to the specific sample. The purpose of optimization is to minimize the discrepancies between the real distribution and the estimated distribution. GTNP uses the optimal estimated distribution as the posterior distribution for the inference of testing.

Finally, ${z}$ and ${u}$ obtained from a single sample are used to obtain the final detection result via a fully connected layer (FC).
\begin{equation}
{y}_{s}, {y}_{t} = FC\left(\left[\mathbf{z}_{s} \mid {u}_{s} \right], \left[\mathbf{z}_{t} \mid {u}_{t}\right] \right)
\label{FC}
\end{equation}
where $[\cdot \mid \cdot]$ means concatenate 2 variables.

\subsection{Embedding}
Embedding is used to construct the latent embedding representations of input data in a latent space. The shared weight feature extractor we employed is a convolutional neural network (CNN). GTNP concatenates the outputs of CNN with the global latent variable ${z_{global}}$.
\begin{equation}
{h}=\left[C N N({x}) \mid {z}_{\text {global }}\right]
 \label{CNN}
\end{equation}
where ${x}$ is the input data, ${h}$ is a vector that contains the information about the data of the input sample and the overall GTNP. We use equation \ref{CNN} to build a unique distribution for each sample and assume that each distribution is independent. Therefore, the distribution of the entire domain can be represented as:
\begin{equation}
p_{\varphi}({U} \mid {H})=\prod_{i \in \text { domain }} p_{\varphi_{i}}\left({u}_{{i}} \mid {h}_{{i}}\right)
 \label{PU}
\end{equation}
where the domain includes the source domain and target domain. ${\varphi_{i}}$ and ${u_i}$ represent the distribution parameters and the latent representation corresponding to sample ${x_i}$, respectively. ${U}$ and ${H}$ represent the set of ${u_i}$ and ${h}_{{i}}$.

\subsection{GraphConstruction}
\begin{figure*}[!ht]
\centering
\includegraphics[width=5.5in]{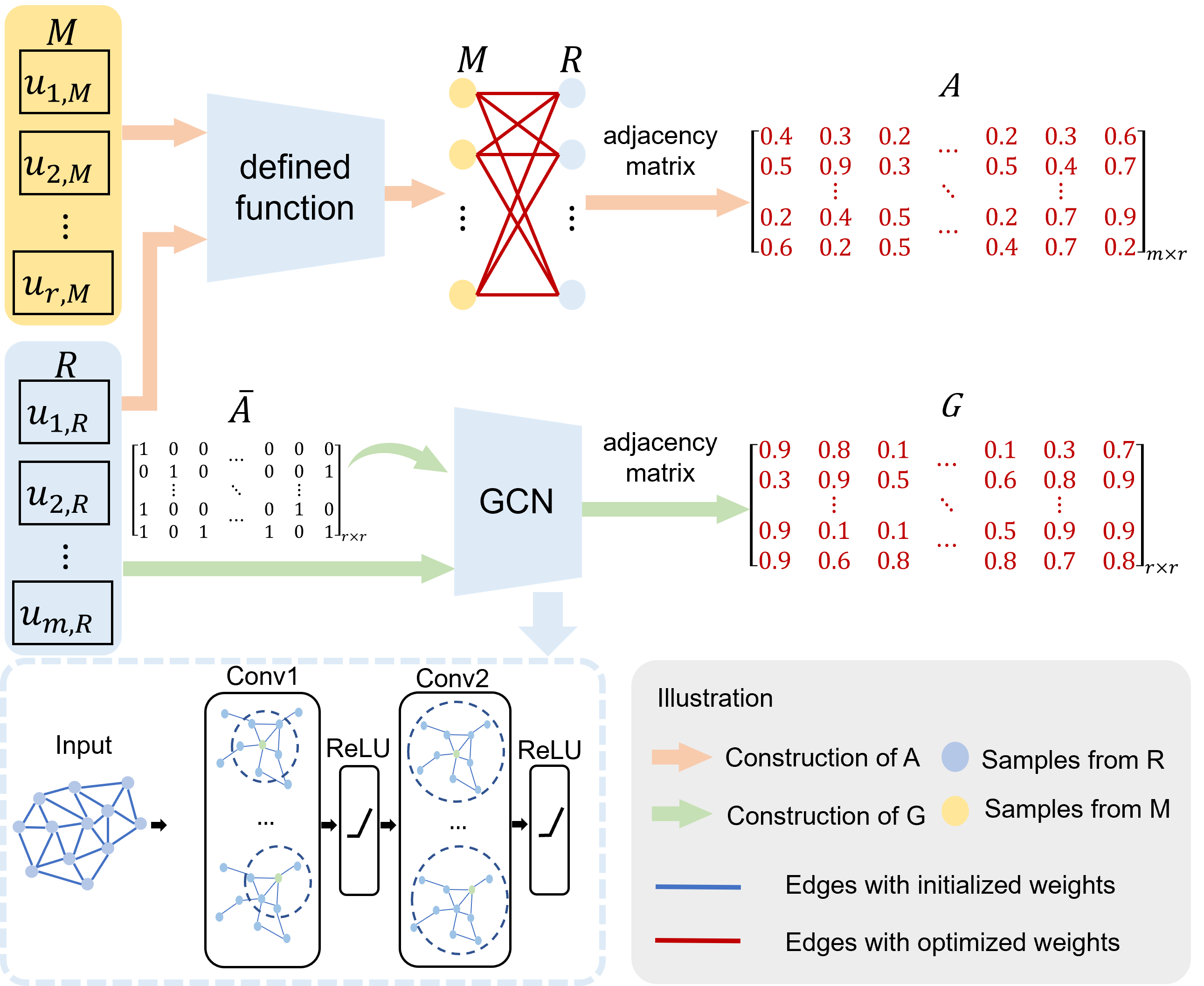}
\caption{The calculation process of the dependency graphs G and A.}
\label{fig_13}
\end{figure*}

In GTNP, two dependency graphs (G and A) are constructed to replace the role of the kernel function in GPs to encode the correlation structure between 2 inputs. G is the adjacency matrix of a weighted undirected graph which is used to represent the relationship of samples in the R set, and A is the adjacency matrix of a bipartite graph which is used to represent the relationship of samples between the R set and the M set, as shown in the figure \ref{fig_13}.

GCN is a deep learning method specifically designed for analyzing graph-structured data. We use samples in the R set to construct an input graph of GCN, where the nodes are samples and the weights in edges are relative relationships between samples. Partial labels of the samples in R set are employed to construct the weights on the edges of the input graph (assigned as 1 if they belong to the same class and 0 otherwise). Each convolution layer in GCN enables the node to gather information from its neighbors and then update its own information based on what it has gathered. The formula for updating is shown in equation \ref{UL}.
\begin{equation}
{u}^{(l+\mathbf{1})}=\sigma\left(\bar{A} {u}^{(l)} W^{(l)}+{b}^{(l)}\right)
 \label{UL}
\end{equation}
where $l$ represents the number of layers of the GCN network, ${u}^{(l)}$ represents the node representation of the $l^{th}$ layer and $\bar{A}$ represents the adjacency matrix of the input to GCN. After the calculation of the convolutional layer, the feature vectors of the 2 nodes are concatenated as the representation of the edge connecting the two nodes.
\begin{equation}
{e}_{i, j}=\left[{u}_{i} \mid {u}_{j}\right]
 \label{EU}
\end{equation}

Taking ${e}_{i, j}$ as the input through a fully connected (FC) layer to obtain the weight corresponding to this edge:
\begin{equation}
w_{i, j}=F C\left({e}_{i, j}\right)
 \label{W}
\end{equation}

In A, edges only exist between samples from different sets. The weights in edges are calculated by equation \ref{G}.
\begin{equation}
w\left({u}_{i}, {u}_{j}\right)=\exp \left(-\frac{\tau}{2}\left\|{u}_{i}-{u}_{j}\right\|^{2}\right)
 \label{G}
\end{equation}

Equation \ref{G} is similar to the radial basis function (RBF) in GPs. $\tau$ is optimized to find a suitable approximate distribution for parameter estimation and prediction in the inference of GTNP.

\subsection{Distribution}
The real distribution and the estimated distribution of samples in the domains are constructed, respectively. In order to use a few samples to construct the real distribution, which guides GTNP to model decent sample distribution in domains, we split the domain into 2 sets. However, since the samples used by G only provide information about R set, we introduce the information from other samples via A to make the constructed real distribution more accurate. Similarly, we suppose that the distribution of each sample is unique and assume that each distribution is independent. The real distribution is shown in equation \ref{P_theta}.
\begin{equation}
\begin{array}{c}
{p}_{\theta_{r}}\left({Z} \mid G, A, {U}_{{R}}, {Y}_{{R}}\right)=\prod_{i \in R \cup M} {p}_{\theta_{r, i}}\left(\mathbf{z}_{{i}} \mid G, A, {u}_{{R}, {i}}, {y}_{{R}, {i}}\right) \\
=\prod_{j \in R} {p}_{\theta_{r, j}}\left({z}_{{j}} \mid a d j_{G_{j}}\left({u}_{{R}}, {y}_{{R}}\right)\right) \prod_{k \in M} {p}_{\theta_{r, k}}\left(\mathbf{z}_{{k}} \mid a d j_{A_{k}}\left({u}_{{R}}, {y}_{{R}}\right)\right)
\end{array}
 \label{P_theta}
\end{equation}
where $a d j_{G_{j}}(\cdot)$ and $a d j_{A_{k}}(\cdot)$ are the functions that return the adjacent node of the node $j$, $k$ according to G, A respectively. ${y_R}$ represents the label of the samples in R.

Having obtained the real distribution, we are now interested in how to fit the parameters $\theta_{r}$ when GTNP gets the new input ${x}$ during the process of testing. However, the real distribution is difficult to calculate accurately for complex models. VI provides an alternative method by finding a simple distribution that approximates the real distribution. Specifically, VI is a method by defining an estimated distribution and then adjusting the parameters of this distribution to make it as close as possible to the real distribution. The estimated distribution is constructed by equation \ref{q_theta} based on latent embedding representations ${u}$.
\begin{equation}
{q}_{\theta_{e}}({Z} \mid {U})=\prod_{i \in R \cup M} {q}_{\theta_{e}}\left(\mathbf{z}_{{i}} \mid {u}_{{i}}\right)
 \label{q_theta}
\end{equation}

The parameters $\theta_{e}$ of the estimated distribution are trained by minimizing the discrepancy (typically measured by the Kullback-Leibler (KL) divergence) between the real and estimated distributions.

\subsection{Loss Function}
The loss function of GTNP is expressed as follows:
\begin{equation}
 Loss =\overbrace{L_{s_{distribution}} + L_{t_{distribution}}}^{L_{ {distribution}}}+\overbrace{L_{s_{classification}} + L_{t_{classification}}}^{L _ {classification}}+L_{{MMD}}
 \label{loss_all}
\end{equation}
$L_{distribution}$ and $L_{classification}$ include the loss both on the source domain and the target domain, and the calculation processes for 2 domains are the same. This section takes the loss calculation on a single domain as an example. The 3 loss terms are explained as follows:

(1) $L_{ {distribution }}$ is used to minimize the discrepancy between ${p}_{\theta_{r}}$ and ${q}_{\theta_{e}}$. KL divergence is used to describe the discrepancy between 2 distributions. Therefore, GTNP has the following optimization problem:
\begin{equation}
{q}^{\star}_{\theta_{e}}\left(Z \mid {U}\right)=\arg \min _{q({z}) \in \mathcal{Q}} D_{\mathrm{KL}}[{q}_{\theta_{e}}\left(Z \mid {U}\right) \| {p}_{\theta_{r}}\left({Z} \mid G, A, {U}_{{R}}, {Y}_{{R}}\right)]
 \label{qz_star}
\end{equation}
where the KL divergence is written as:
\begin{equation}
D_{\mathrm{KL}}[{q}_{\theta_{e}}\left(Z \mid {U}\right) \| {p}_{\theta_{r}}\left({Z} \mid G, A, {U}_{{R}}, {Y}_{{R}}\right)]=\\
-\int_{{Z}} {q}_{\theta_{e}}\left(Z \mid {U}\right) \log \left[\frac{{p}_{\theta_{r}}\left({Z} \mid G, A, {U}_{{R}}, {Y}_{{R}}\right)}{{q}_{\theta_{e}}\left(Z \mid {U}\right)}\right] d {Z}
 \label{kl}
\end{equation}

However, the real distribution ${p}_{\theta_{r}}$ is difficult to calculate, so the KL divergence is expanded in the mathematical expectations form, while expanding the ${p}_{\theta_{r}}$ according to the conditional probability formula:
\begin{equation}
\begin{array}{l}
D_{\mathrm{KL}}[{q}_{\theta_{e}}\left(Z \mid {U}\right) \| {p}_{\theta_{r}}\left({Z} \mid G, A, {U}_{{R}}, {Y}_{{R}}\right)] \\
=\mathbb{E}_{Z \sim {q}_{\theta_{e}}}[\log {q}_{\theta_{e}}\left(Z \mid {U}\right)]-\mathbb{E}_{Z \sim {q}_{\theta_{e}}}[\log {p}_{\theta_{r}}\left({Z} \mid G, A, {U}_{{R}}, {Y}_{{R}}\right)] \\
=\mathbb{E}_{Z \sim {q}_{\theta_{e}}}[\log {q}_{\theta_{e}}\left(Z \mid {U}\right)]-\mathbb{E}_{Z \sim {q}_{\theta_{e}}}[\log {p}_{\theta_{r}}\left({Z}, G, A, {U}_{{R}}, {Y}_{{R}}\right)]+\mathbb{E}_{Z \sim {q}_{\theta_{e}}}[\log {p}_{\theta_{r}}\left({Z} \mid G, A, {U}_{{R}}, {Y}_{{R}}\right)] \\
=\mathbb{E}_{Z \sim {q}_{\theta_{e}}}[\log {q}_{\theta_{e}}\left(Z \mid {U}\right)]-\mathbb{E}_{Z \sim {q}_{\theta_{e}}}[\log {p}_{\theta_{r}}\left({Z}, G, A, {U}_{{R}}, {Y}_{{R}}\right)]+\log {p}_{\theta_{r}}\left(G, A, {U}_{{R}}, {Y}_{{R}}\right)
\end{array}
 \label{ekl}
\end{equation}

Taking equation \ref{ekl} into equation \ref{qz_star}, which can be rewritten as:
\begin{equation}
\begin{aligned}
{q}^{\star}_{\theta_{e}}\left(Z \mid {U}\right) & =\arg \max _{{q}_{\theta_{e}}\left(Z \mid {U}\right) \in \mathcal{Q}} \mathbb{E}_{Z \sim q_{\theta_{e}}}\left[\log {p}_{\theta_{r}}\left({Z}, G, A, {U}_{{R}}, {Y}_{{R}}\right)-\mathbb{E}_{Z \sim q_{\theta_{e}}}[\log {q}_{\theta_{e}}\left(Z \mid {U}\right)]\right] \\
& =\arg \max _{{q}_{\theta_{e}}\left(Z \mid {U}\right) \in \mathcal{Q}} \operatorname{ELBO}(q_{\theta_{e}})
\end{aligned}
 \label{q_star_new}
\end{equation}
where ELBO is called Evidence Lower Bound, which is the lower bound of the marginal likelihood. $L_{distribution}$ is defined in equation \ref{l_dis}, so that minimizing $L_{distribution}$ is equivalent to maximizing ELBO. The optimal estimated parameters $\theta_{e}$ are found by maximizing ELBO, which means making the estimated distribution as close as possible to the real distribution. More details of optimizing ELBO refer to \textbf{section 2.2} in the Appendix.
\begin{equation}
{L}_{{distribution }}=-\mathbb{E}_{Z \sim q_{\theta_{e}}}\left[\log {p}_{\theta_{r}}\left({Z}, G, A, {U}_{{R}}, {Y}_{{R}}\right)-\mathbb{E}_{Z \sim q_{\theta_{e}}}[\log {q}_{\theta_{e}}\left({Z} \mid {U}\right)]\right]
 \label{l_dis}
\end{equation}

(2) $L_{{classification}}$ is the loss of classification between the model’s predictions and the labels. Since the outputs of GTNP are predicted probability values, CrossEntropyLoss is choosed to calculate the loss between the predictions and the labels.
\begin{equation}
L \left._{{classification}}={CrossEntropyLoss (outputs, labels}\right)=-\frac{1}{N} \sum_{i=1}^{N} \sum_{c=1}^{M} y_{i c} \log \left(p_{i c}\right)
 \label{l_classification}
\end{equation}
where $N$ is the number of the samples, $M$ is the the number of the classes, $y_{ic}$ is a sign function that takes 1 if the true class of sample $i$ is $c$ and 0 otherwise and $p_{ic}$ is the predicted probability value that sample $i$ belongs to $c$.

(3) $L_{\text{MMD}}$ is the distribution discrepancies of samples between the source domain and target domain. Due to the complex distribution of samples, the distribution function cannot be directly given. Therefore, we regard the samples as random variables and use the moments of the random variables to describe the distribution. If 2 random variables have more moments of different order that are similar, it means that the distributions of these 2 random variables are closer. MMD is used to calculate the discrepancy between the distributions of 2 domains, which use the moment with the largest discrepancy between 2 distributions as the standard to measure the 2 distributions \cite{li2020intelligent}. The calculation of MMD is defined in equation \ref{mmd_1}.
\begin{equation}
\begin{aligned}
L_{{MMD}} & =\left\|\frac{1}{n_{s}} \sum_{i=1}^{n_{s}} f\left(u_{i}\right)-\frac{1}{n_{t}} \sum_{j=1}^{n_{t}} f\left(u_{j}\right)\right\|_{\mathcal{H}}^{2} \\
& =\left\|\frac{1}{{n_{s}}^{2}} \sum_{i=1}^{n_{s}} \sum_{i^{\prime}=1}^{n_{s}} k\left(u_{i}, u_{i}^{\prime}\right)-\frac{2}{n_{s} n_{t}} \sum_{i=1}^{n_{s}} \sum_{j=1}^{s_{t}} k\left(u_{i}, u_{j}\right)+\frac{1}{{n_{t}}^{2}} \sum_{j=1}^{n_{t}} \sum_{j^{\prime}=1}^{n_{t}} k\left(u_{j}, u_{j}^{\prime}\right)\right\| \\
& =\operatorname{tr}\left(\left[\begin{array}{ll}
K_{s, s} & K_{s, t} \\
K_{t, s} & K_{t, t}
\end{array}\right] {M}\right)
\end{aligned}
 \label{mmd_1}
\end{equation}
where
\begin{equation}
(M)_{i j}=\left\{\begin{array}{ll}
\frac{1}{n_{s} n_{s}}, & {u}_{i}, {u}_{j} \in \mathcal{D}_{s} \\
\frac{1}{n_{t} n_{t}}, & {u}_{i}, {u}_{j} \in \mathcal{D}_{t} \\
\frac{-1}{n_{s} n_{t}}, & \text { otherwise }
\end{array}\right.
 \label{mmd_2}
\end{equation}
where $f(\cdot)$ represents mapping function, and $k(\cdot)$ represents gaussian kernel function. The $K_{s, s}$, $K_{s, t}$, $K_{t, s}$, $K_{t, t}$ are kernel matrices (also called Gram matrices). To speed up the training process, the Adam algorithm \cite{kingma2014adam} is used to optimize equation \ref{loss_all}.

\section{Experiments and analysis} \label{sec:results}
\subsection{Dataset and preprocessing}
For the bearing fault detection task, we use the Case Western Reserve University (CWRU) rolling bearing dataset \cite{hendriks2022towards}, which encompasses diverse operational conditions and employs the SKF rolling bearing model 6205-2RS. In this dataset, vibration signals from rolling bearing are collected at a 48 $k$Hz sampling frequency, covering 4 distinct working conditions with rotational speeds of $1797$, $1772$, $1750$, and $1730$ RPM (denoted as C0, C1, C2 and C3), respectively. In each working condition, single-point faults are induced in the ball, inner raceway, and outer raceway, with fault diameters of 0.178 $mm$, 0.356 $mm$ and 0.532 $mm$, respectively, resulting in $9$ fault types and $1$ normal type. To enrich the feature representation extracted by GTNP, we employ non-overlapping sliding window segmentation to process the original vibration signals in the CWRU dataset. Specifically, each image ($32 \times 32$) comprises $1024$ non-overlapping vibration data samples.

In addition, in order to verify the performance of the proposed method in complex tasks, we collect measurement data from simulated and real aircraft that cover 3 simulated aircraft and 1 real aircraft, which provide data for cross-machines and emerging fault detection tasks. This dataset covers a variety of aircraft modes, including high and low altitudes for both cruise and manual free flight. Each aircraft has 9 fault types (including excursions, oscillations, drifts,  etc) and 1 normal type. We normalize the measurement data and stack it into a two-dimensional matrix. In this matrix, each row represents data sequences from 15 sensors, covering 12 aircraft states and 3 load coefficients, while each column corresponds to the data from each sensor at the measurement moment. It is worth noting that since the data are downsampled to a frequency of 1 Hz (the time window is 30 seconds), the matrix contains $31$ columns. The final size of each matrix is $15 \times 31$ after preprocessing. \textbf{Section 1.1} of Appendix provides a more detailed description of the datasets and preprocessing.

\subsection{Evaluation metrics}
In this work, a total of 6 evaluation metrics are considered to evaluate the performance of the proposed methods, including accuracy ($acc$), precision ($prec$), recall ($rec$), f1-score ($f1$), receiver operating characteristic curve ($roc$) and area under curve ($auc$). Utilizing evaluation metrics from the binary classification tasks, we have formulated the calculation formulas of metrics for the multi-classification tasks involved in this paper. \textbf{Section 1.2.1} gives a detailed calculation process for the 6 evaluation metrics in the Appendix.

\subsection{Method configurations}
All experiments are conducted and developed using the PyTorch 2.0.0+cu117. The processes of code debugging and the visualization of results are executed with the PyCharm 2021.2 (Professional Edition) and Jupyter Notebook 1.0.0. For the optimization of GTNP, we adopt the RMSprop optimizer with a learning rate set to 0.01 for the CWRU dataset and the Adam optimizer with a learning rate set to 0.01 for the aircraft dataset, respectively. For a more comprehensive understanding of the experimental setup, please refer to \textbf{Table 6} in the Appendix.

\subsection{Case study of bearing fault detection with cross-working conditions}
\subsubsection{Construct the representation of source domain knowledge}
GCN is introduced to enhance the classification performance of GTNP in the target domain by utilizing detection knowledge learned from the source domain \cite{song2021scgcn}. For CWRU dataset, we randomly select 300 samples from the source domain to construct a graph, where the nodes and the weighted edges of the graph represent samples and the dependency between samples, respectively. The primary function of the GCN is to model the dependencies between these 300 samples, essentially obtaining the weights of edges. The output of trained GCN is a graph with optimal weights of edges, which is utilized as prior knowledge for the training of GTNP. This section focuses on 2 crucial hyperparameters: the learning rate and the optimizer. We test their impact on the model's performance. The training results on CWRU dataset are illustrated in Figure \ref{fig_1}.

For the learning rate, Figure \ref{fig_1} shows that a learning rate of 0.001 hinders the optimization process and has an adverse impact on accuracy convergence. Additionally, the choice of optimizer significantly influences detection accuracy. Comparing with ASGD, RMSprop, and ADAM, it is clear that RMSprop outperforms the others. RMSprop enhances the stability of the training by adaptively adjusting the learning rate to meet the specific requirements of each parameter. Moreover, RMSprop demonstrates increased robustness to sparse data within the CWRU dataset, as it automatically fine-tunes the learning rate for different parameters, leading to varying learning rates across distinct features. Therefore, RMSprop and the learning rate of 0.01 are fixed for experiments on CWRU dataset.

\begin{figure*}[!h] 
    \centering
    \begin{subfigure}[b]{0.42\textwidth}
        \includegraphics[width=\textwidth]{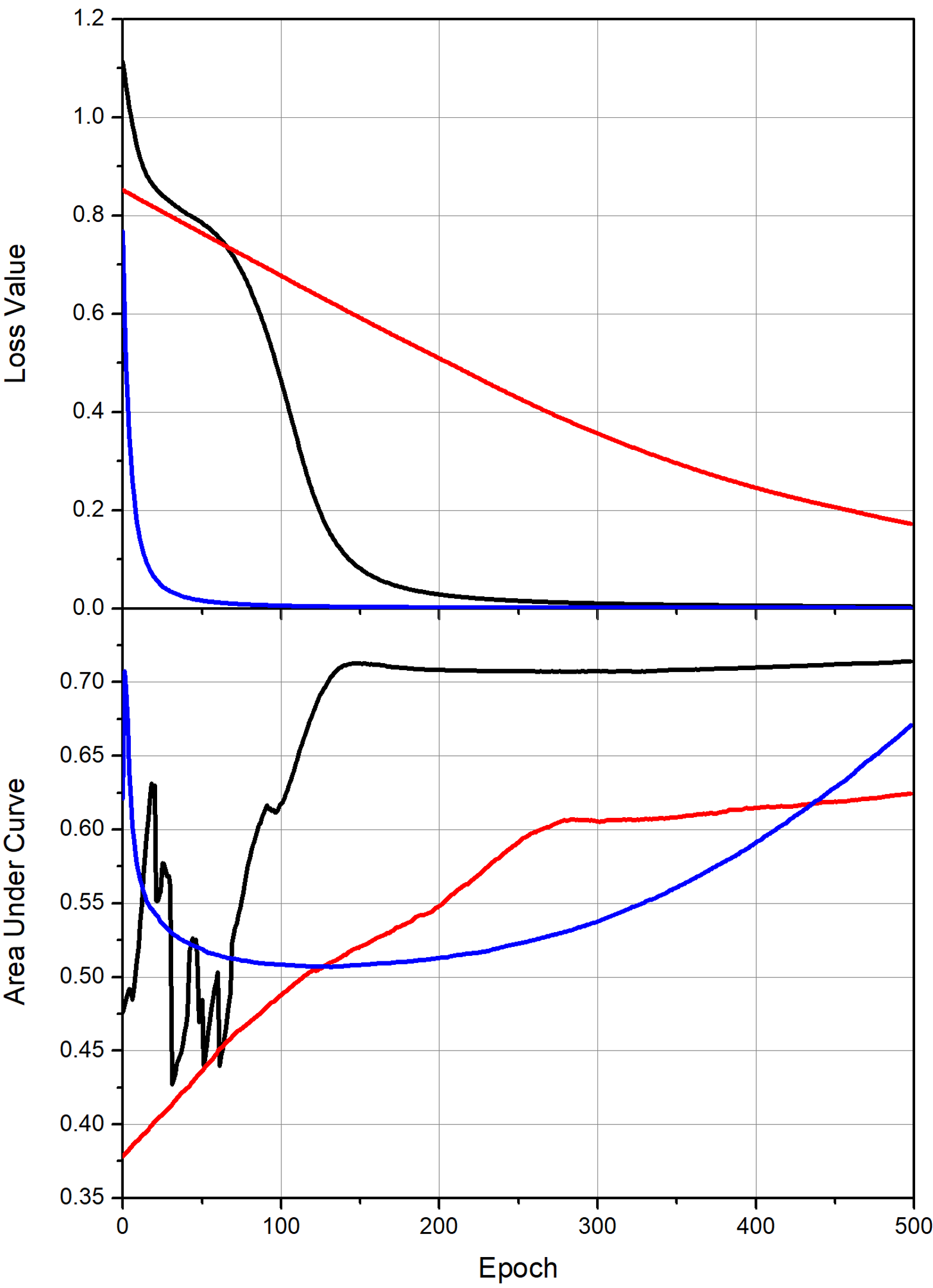}
        \caption{}
    \end{subfigure}
    \begin{subfigure}[b]{0.528\textwidth}
        \includegraphics[width=\textwidth]{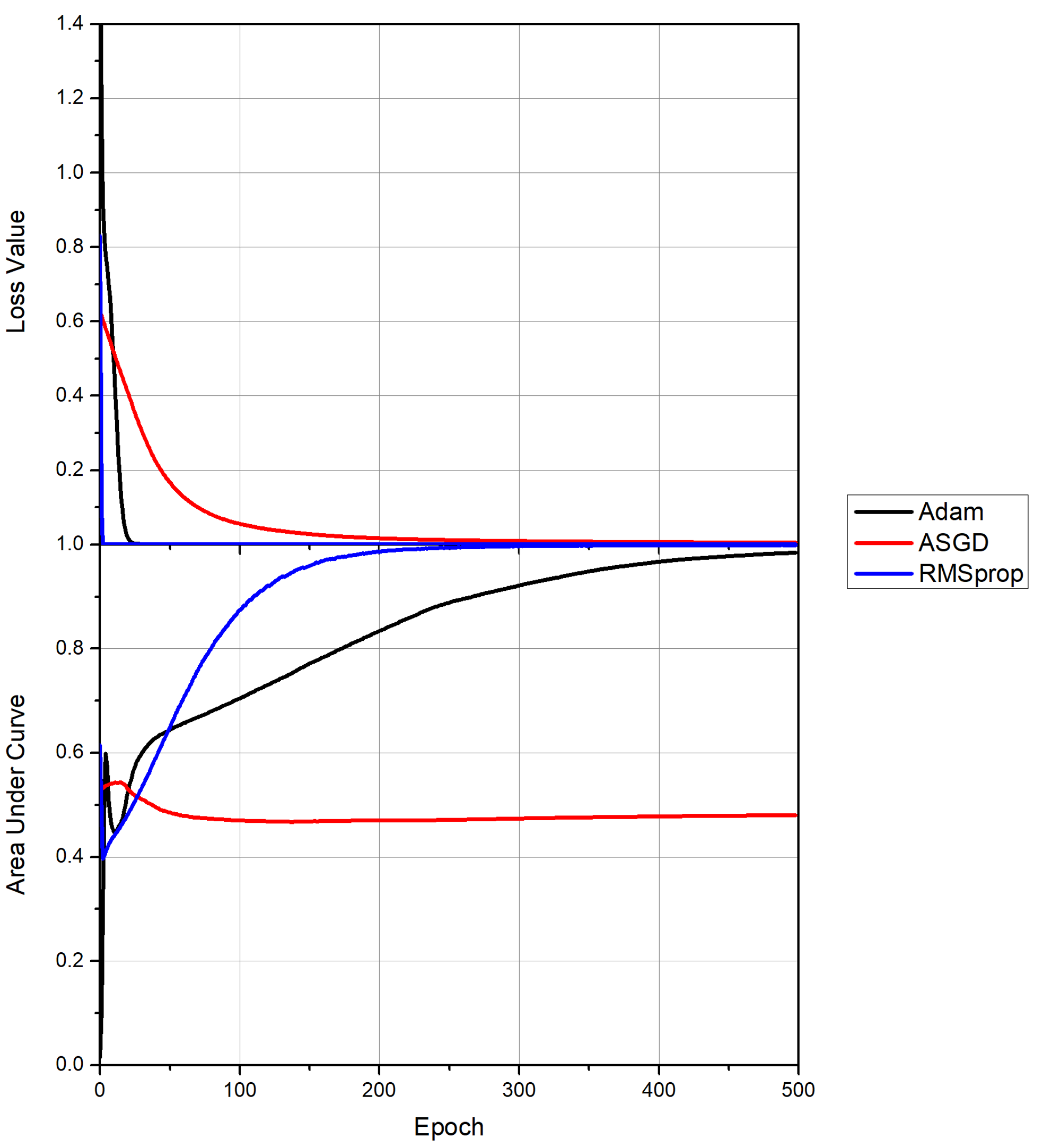}
        \caption{}
    \end{subfigure}
    \caption{GCN training results on CWRU dataset. (a) The learning rate is set to 0.001, 
    (b) the learning rate is set to 0.01.}
    \label{fig_1}
\end{figure*}

\subsubsection{Analysis of hyperparameter settings for GTNP training}
\begin{figure}[!h]
    \centering
    \subfloat[batch Size 32]{\includegraphics[width=0.45\textwidth]{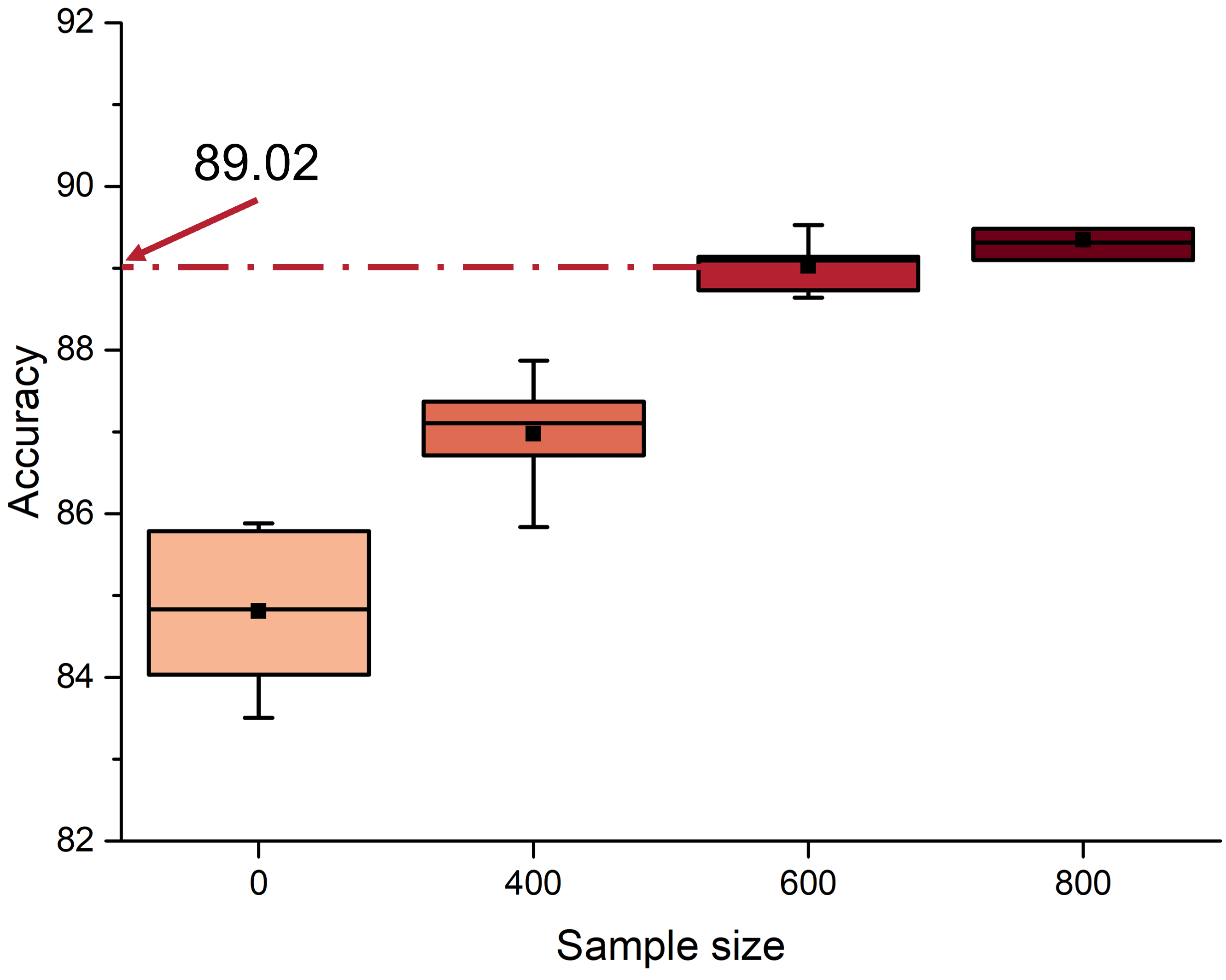}}
    \hspace{0.02\textwidth}
    \subfloat[batch Size 64]{\includegraphics[width=0.45\textwidth]{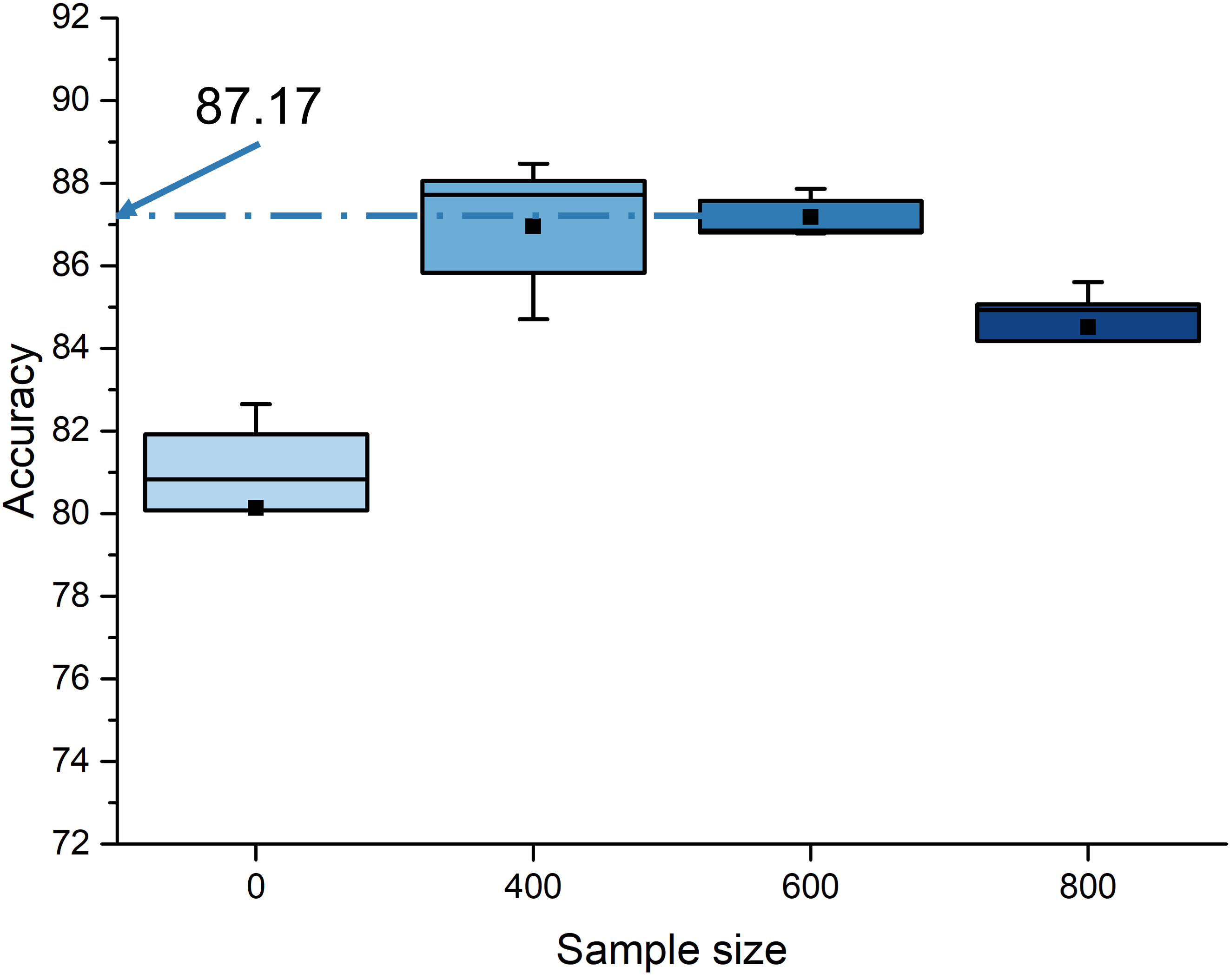}}\\
    \subfloat[batch Size 128]{\includegraphics[width=0.45\textwidth]{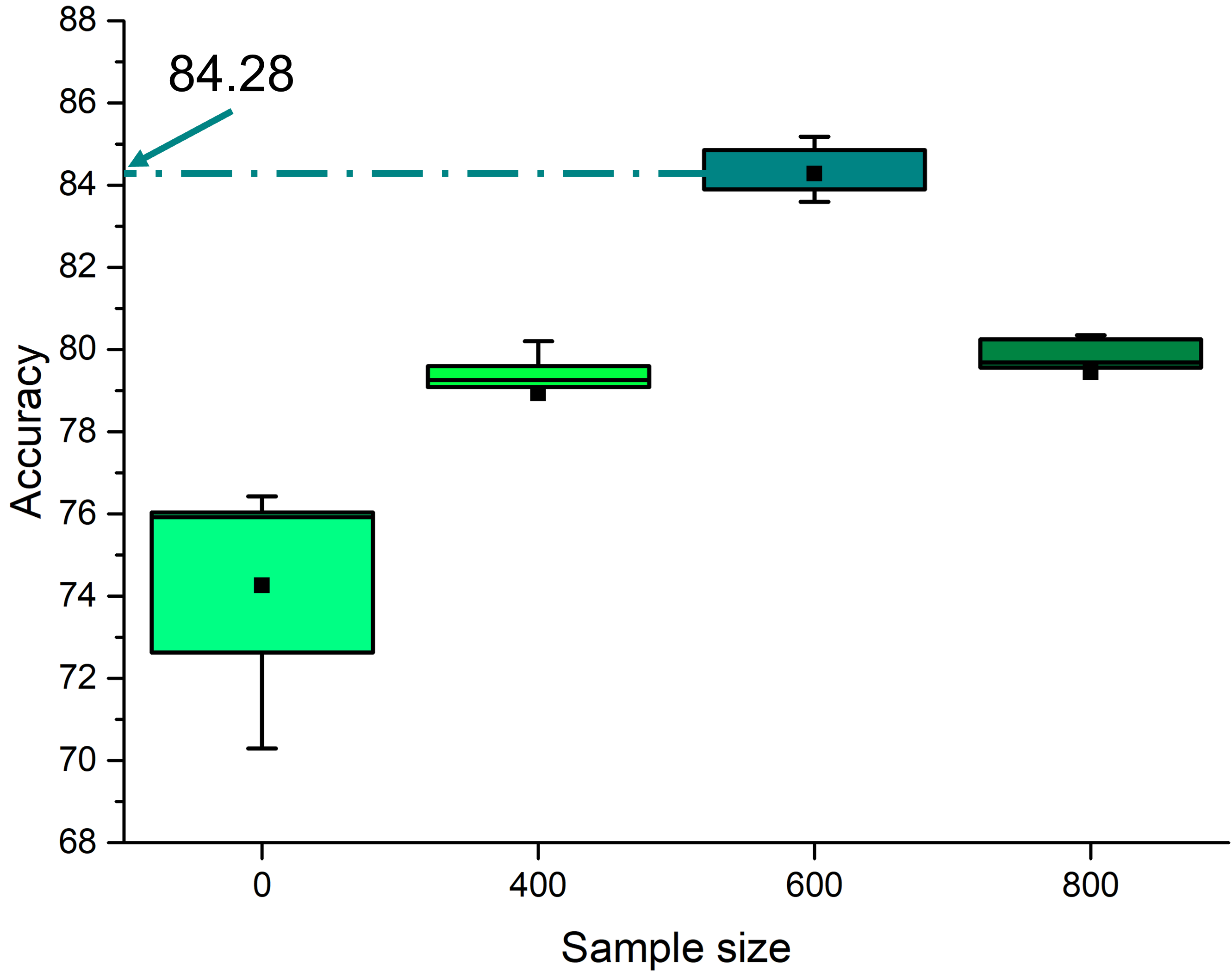}}
    \caption{Box plot of comparison results of batch size and sample size.}
    \label{fig_2}
\end{figure}

The purpose of the reference set is to mitigate the demands of sample size utilized in the target domain by the GTNP. To evaluate how the sample size of the reference set impacts the performance of the GTNP, we use data in working condition C0 as the source domain and data in working condition C1 as the target domain. In the comparison experiments, we select different sample sizes of reference set in target domain, specifically 0, 400, 600, and 800. Additionally, 3 sets of experiments are designed with different batch sizes (32, 64, and 128) during training. Each experiment is repeated 10 times, and we calculate the mean and standard deviation of model accuracy. The classification results in the target domain are presented in Figure \ref{fig_2} and Table \ref{table_3}.

\begin{table}[!ht]
    \centering
    \caption{Comparison experiments of batch size and sample size of the reference set.}
    \begin{tabular}{c|cc|cc|cc}
    \hline
        Batch size & \multicolumn{2}{|c|}{32} & \multicolumn{2}{|c|}{64} & \multicolumn{2}{|c}{128} \\ \hline
        Sample size & 600 & 800 & 600 & 800 & 600 & 800 \\ \hline
        Average (acc) & 89.02 & 89.34 & 87.17 & 84.51 & 84.28 & 79.44 \\ \hline
        Standard Deviation (acc) & 0.35 & 0.69 & 0.5 & 1.08 & 0.69 & 1.19 \\ \hline
    \end{tabular}
    \label{table_3}
\end{table}

For the sample size of reference set, increment in the initial offers a certain degree of improvement in the model's classification accuracy. However, as the sample size increases from 600 to 800, the improvement is minimal (batch size is 32) or even decreases (batch is 64 or 128). This phenomenon may be attributed to the relatively smooth and low-noise of the CWRU dataset. Excessive samples in the reference set lead to overfitting, thereby impairing the model's performance in the target domain. Furthermore, with a fixed sample size of 600, comparisons between different batch sizes reveal a decreasing trend in accuracy as the batch size increases. Large batch sizes reduce randomness and raise the risk of the model being trapped in a local minimum. Consequently, for the comparison experiments, we utilize a reference set consisting of 600 samples with a batch size of 32 for training.

\subsubsection{Analysis of approximate inference in training process}
\begin{figure}[!h]
    \centering
    \subfloat[epoch 0]{\includegraphics[width=0.45\textwidth]{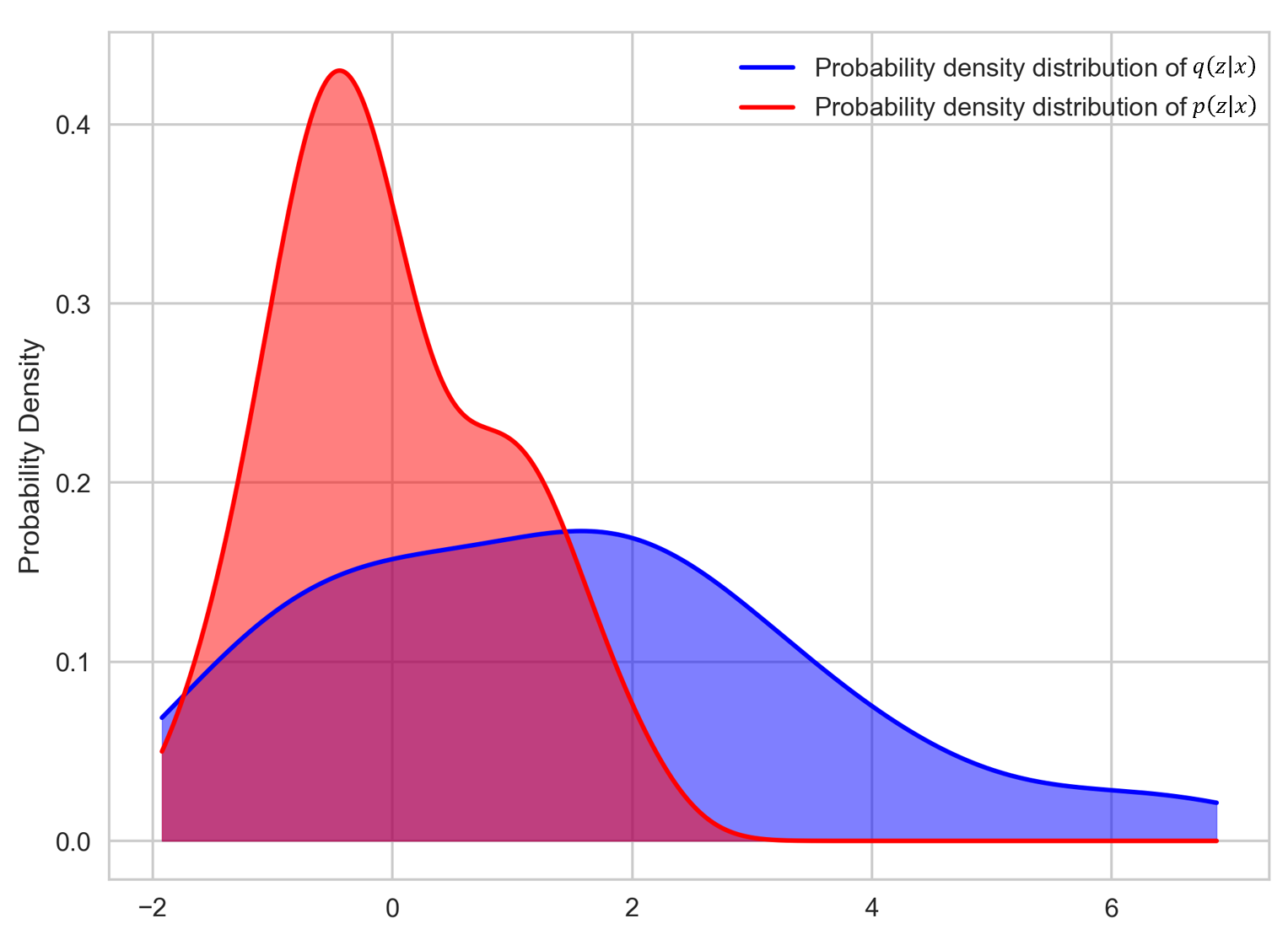}}
    \hspace{0.02\textwidth}
    \subfloat[epoch 5]{\includegraphics[width=0.45\textwidth]{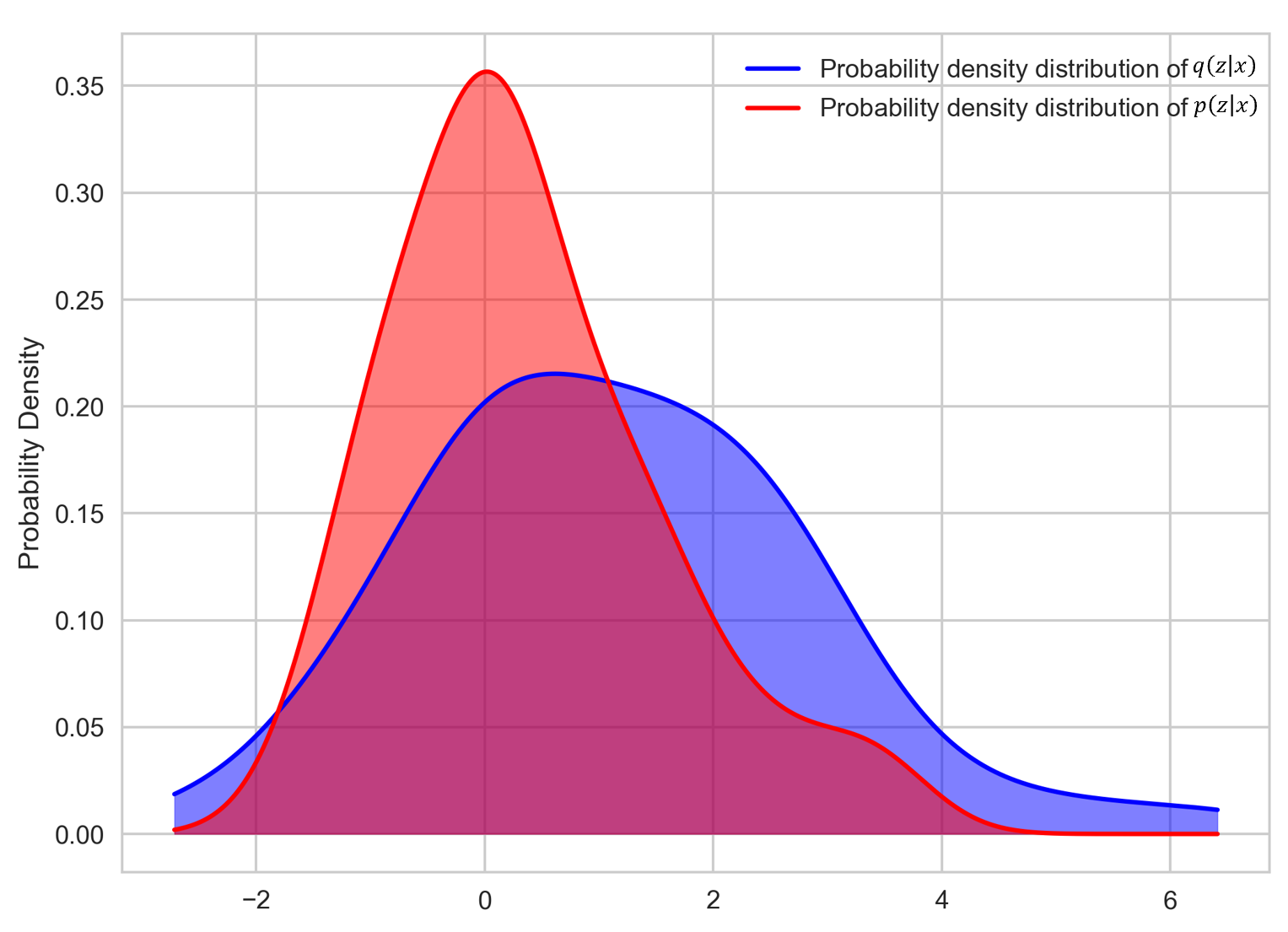}}\\
    \subfloat[epoch 10]{\includegraphics[width=0.45\textwidth]{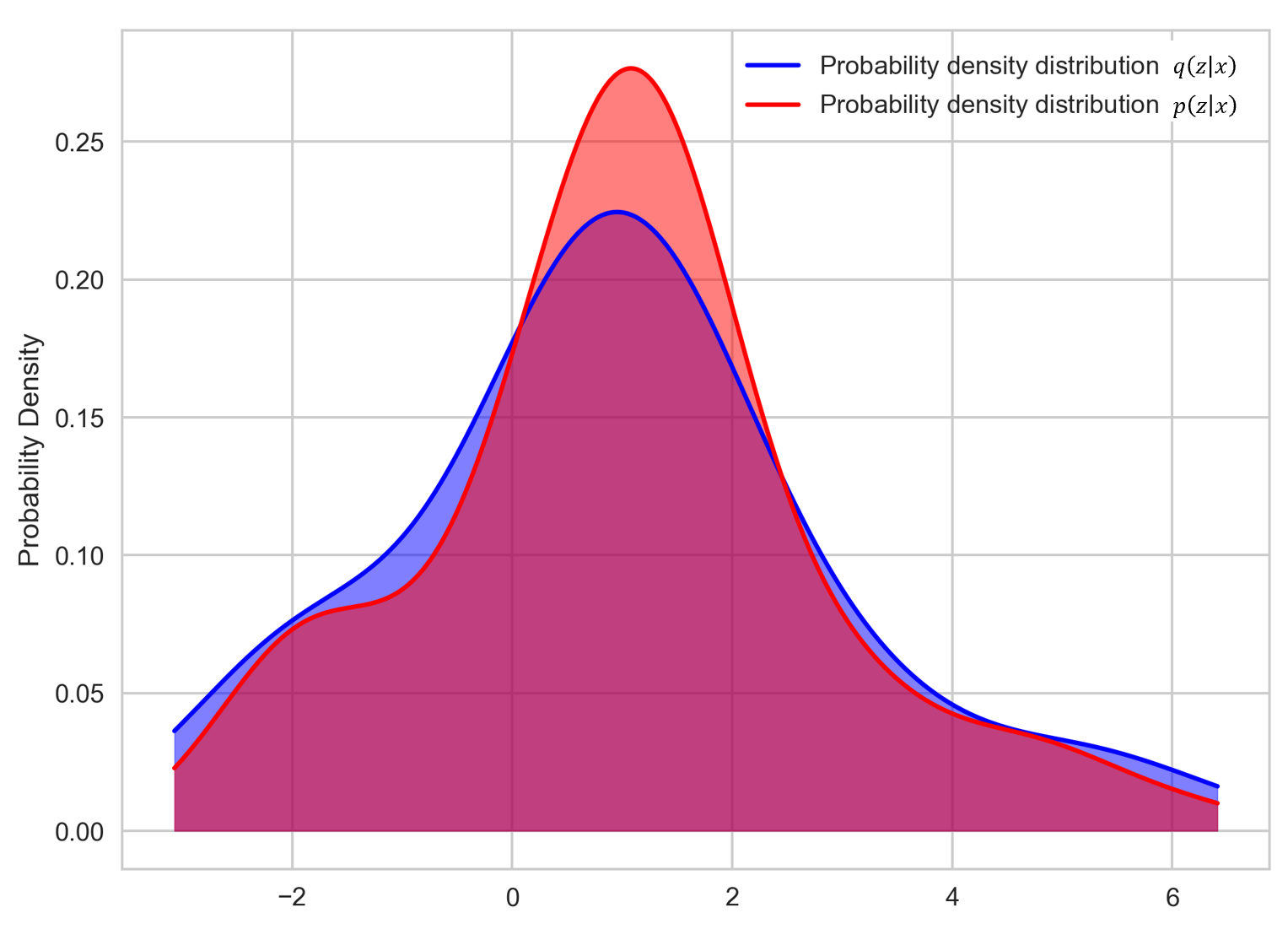}}
    \hspace{0.02\textwidth}
    \subfloat[epoch 20]{\includegraphics[width=0.45\textwidth]{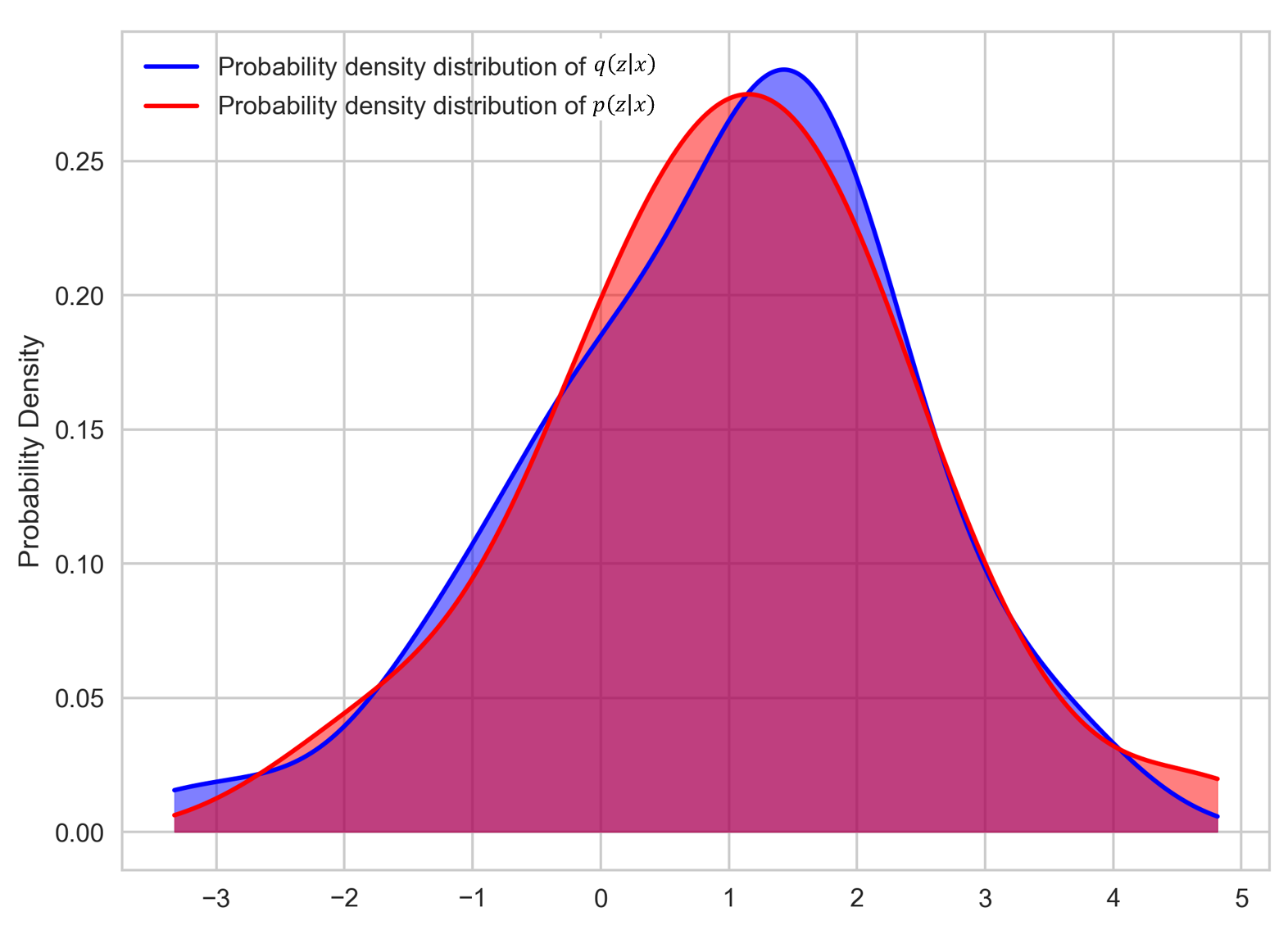}}
    \caption{Probability density estimation of the real posterior distribution $p(z|x)$ and the estimated distribution $q(z|x)$.}
    \label{fig_3}
\end{figure}

\begin{figure*}[!h]
\centering
\includegraphics[width=5.5in]{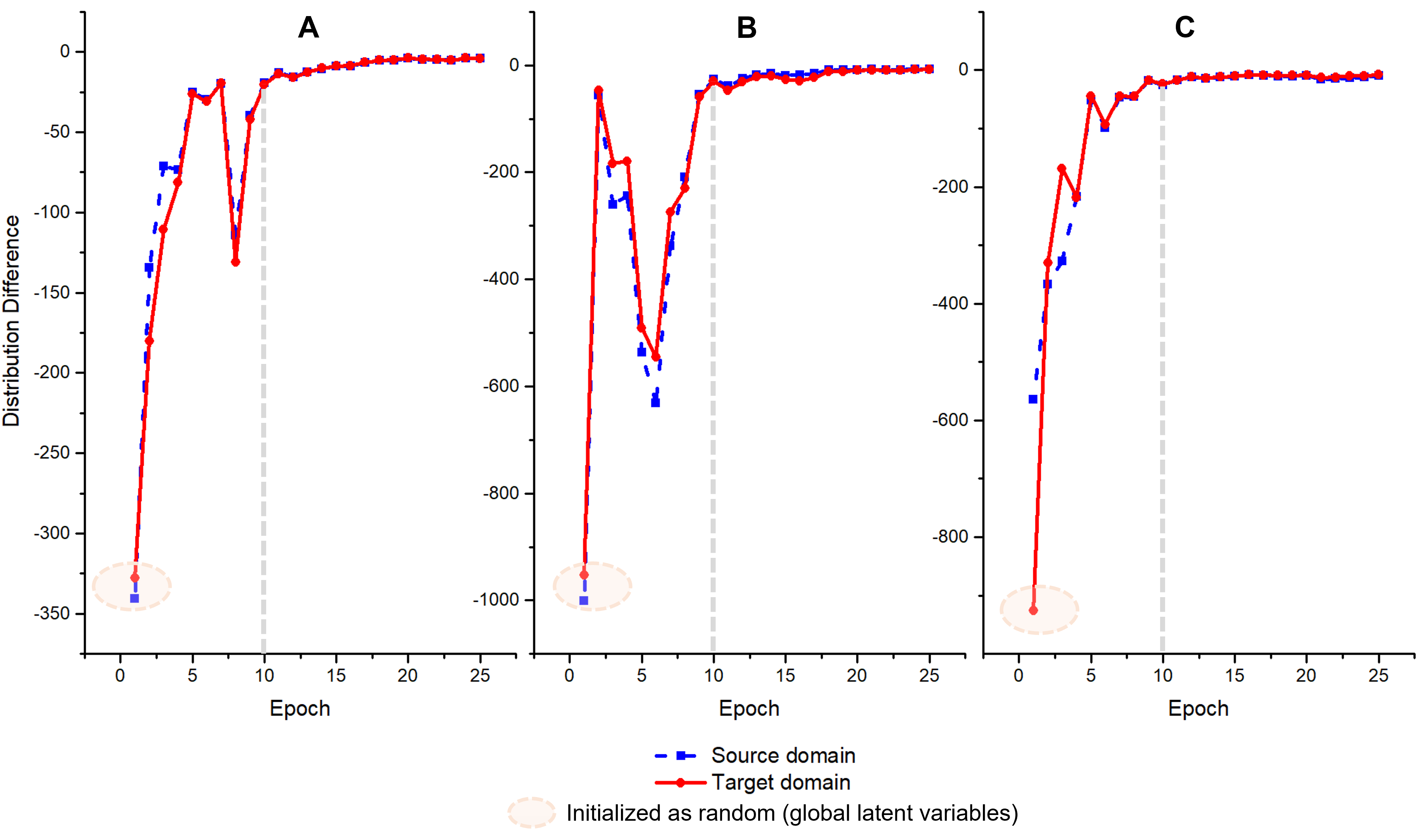}
\caption{Distribution discrepancy of the real posterior distribution $p(z|x)$ and the estimated distribution $q(z|x)$. \textbf{A} Source domain C0 (rotational speeds of 1797 RPM) is transferred to target domain C1 (rotational speeds of 1772 RPM). \textbf{B} Source domain C0 is transferred to target domain C2 (rotational speeds of 1750 RPM). \textbf{C} Source domain C0 is transferred to target domain C3 (rotational speeds of 1730 RPM).}
\label{fig_4}
\end{figure*}

The posterior distribution, obtained during the training, is employed for the inference of GTNP. However, it is often a challenging task to compute the posterior probability distribution analytically. Variational inference (VI) is a widely-adopted technique for the approximate inference of probabilistic method \cite{blei2017variational}. VI aims to identify an estimated distribution, denoted as $q(z|x)$, which closely approximates the real posterior distribution, denoted as $p(z|x)$, containing the local latent variable $z$. To illustrate the impact of how the estimated distribution $q(z|x)$ converges to the real posterior distribution $p(z|x)$, we initially generate 64-dimensional sample vectors from both $p(z|x)$ and $q(z|x)$ during the training process, taking into account their distribution parameters, namely, mean and variance. Subsequently, we fit each distribution using kernel density estimation. The results are shown in Figures \ref{fig_3} and \ref{fig_4}, the discrepancy between the 2 distributions is notable at the beginning of the model training due to the random initialization of the global latent variable. However, as the epoch increases, the discrepancy between the 2 distributions rapidly narrows. When the epoch reaches 10, the discrepancy between the 2 distributions converges. GTNP assess the discrepancy between 2 distributions and leverage that to optimize the parameters of the estimated distribution.

\subsubsection{Comparison with other deep transfer learning methods}
In order to demonstrate the performance of the proposed method, 5 DTL methods are utilized for comparison, which are (1) the well-trained baseline network (ResNet-18) using only data in source domain, (2) the DTL method based on MMD \cite{li2021rolling}, (3) the DTL method based on a mixture of MMD and adversarial training (DANN) \cite{guo2018deep}, (4) the DTL method based on conditional adversarial training (CDANN) \cite{zhang2021conditional} and (5) the NPs-based DTL method (TNP) without GCN proposed in this paper. For a fair comparison, all methods are based on the same training parameters (batch size, optimizer and learning rate). Each experiment is repeated 10 times and the average results are used.

The CWRU dataset includes 4 distinct rotational speeds (specifically, 1797, 1772, 1750, and 1730 RPM), denoted as C0, C1, C2, and C3, respectively. In a specific experiment, we denote $Ci$ as the source domain and $Cj$ ($j \neq i$) as the target domain. Twelve experiments are conducted in total. For the source domain, we divide the training and testing sets employing the 8:2 ratio. As for the target domain, 600 samples that are randomly selected from the target domain construct the training set, which only comprises 12\% of the entire target domain, while the remaining samples are designated as the testing set. In the initial configuration of GTNP, the adaptive amplitude factor is set at 0.1, which constrains model complexity, resembling the regularization techniques in neural networks. Within the context of the target domain, empirical experiments conducted under a single domain demonstrated that GTNP exhibited the highest detection accuracy and the swiftest convergence rate. Detailed results are depicted in \textbf{Figure 8} of the Appendix. The results of the 12 experiments between different domains are tabulated in Table \ref{tab:4}. Notably, the average accuracy of GTNP attains an impressive 90.34\%, a substantial improvement of nearly 10\% over alternative DTL methods, such as MMD and MMD-DANN, and a striking 14.75\% increase relative to the baseline method. Additionally, when compared to TNP, the incorporation of GCN is observable to significantly enhance the construction of detection knowledge in the source domain, yielding positive implications for detection performance. The results demonstrate the notable efficacy of the proposed method on the CWRU dataset.

\begin{table}[!h]
\centering
\caption{The average accuracy ($\%$) on the CWRU dataset with different methods.}
\label{tab:4}
\begin{adjustbox}{width=\textwidth} 
\begin{tabular}{cccccccc}
\hline
Task & ResNet-18 & MMD & MMD-DANN & C-DANN & TNP (Ours) & GTNP (Ours) \\ \hline
C0$\rightarrow$C1 & 78.33& 88.12 & 88.31& 88.95 & 88.14 & \textbf{89.02} \\ \hline
C0$\rightarrow$C2 & 77.4 & 84.58 & 83.45 & 85.76 & 82.32 & \textbf{88.36} \\ \hline
C0$\rightarrow$C3 & 74.48 & 85.22 & 62.21 & 78.98 & 89.64 & \textbf{90.24} \\ \hline
C1$\rightarrow$C0 & 72.81 & 84.47 & 85.16 & 86.34 & 83.9 & \textbf{86.95} \\ \hline
C1$\rightarrow$C2 & 75.52 & 92.76 & 91.17 & 92.27 & 93.34 & \textbf{93.45} \\ \hline
C1$\rightarrow$C3 & 78.44 & 57.67 & 82.34 & 84.67 & 89.42 & \textbf{91.77} \\ \hline
C2$\rightarrow$C0 & 74.06 & 85.56 & 84.24 & 83.73 & 84.49 & \textbf{87.13} \\ \hline
C2$\rightarrow$C1 & 78.75 & 90.43 & 90.3 & 90.86 & 86.32 & \textbf{91.01} \\ \hline
C2$\rightarrow$C3 & 79.98 & 81.75 & 74.65 & 87.73 & 92.32 & \textbf{93.32} \\ \hline
C3$\rightarrow$C0 & 66.77 & 78.34 & 80.38 & 74.12 & 85.13 & \textbf{87.13} \\ \hline
C3$\rightarrow$C1 & 74.38 & 82.91 & 78.32 & 82.92 & 80.29 & \textbf{91.75} \\ \hline
C3$\rightarrow$C2 & 76.15 & 75.63 & 79.78 & 79.84 & 92.94 & \textbf{93.91} \\ \hline
Average & 75.59 & 82.29 & 81.69 & 84.65 & 87.35 & \textbf{90.34} \\ \hline
\end{tabular}
\end{adjustbox}
\end{table}

\subsubsection{Analysis of multi-scale uncertainties of GTNP}
\begin{figure}[!htbp]
    \centering
    \subfloat[]{\includegraphics[width=0.45\textwidth]{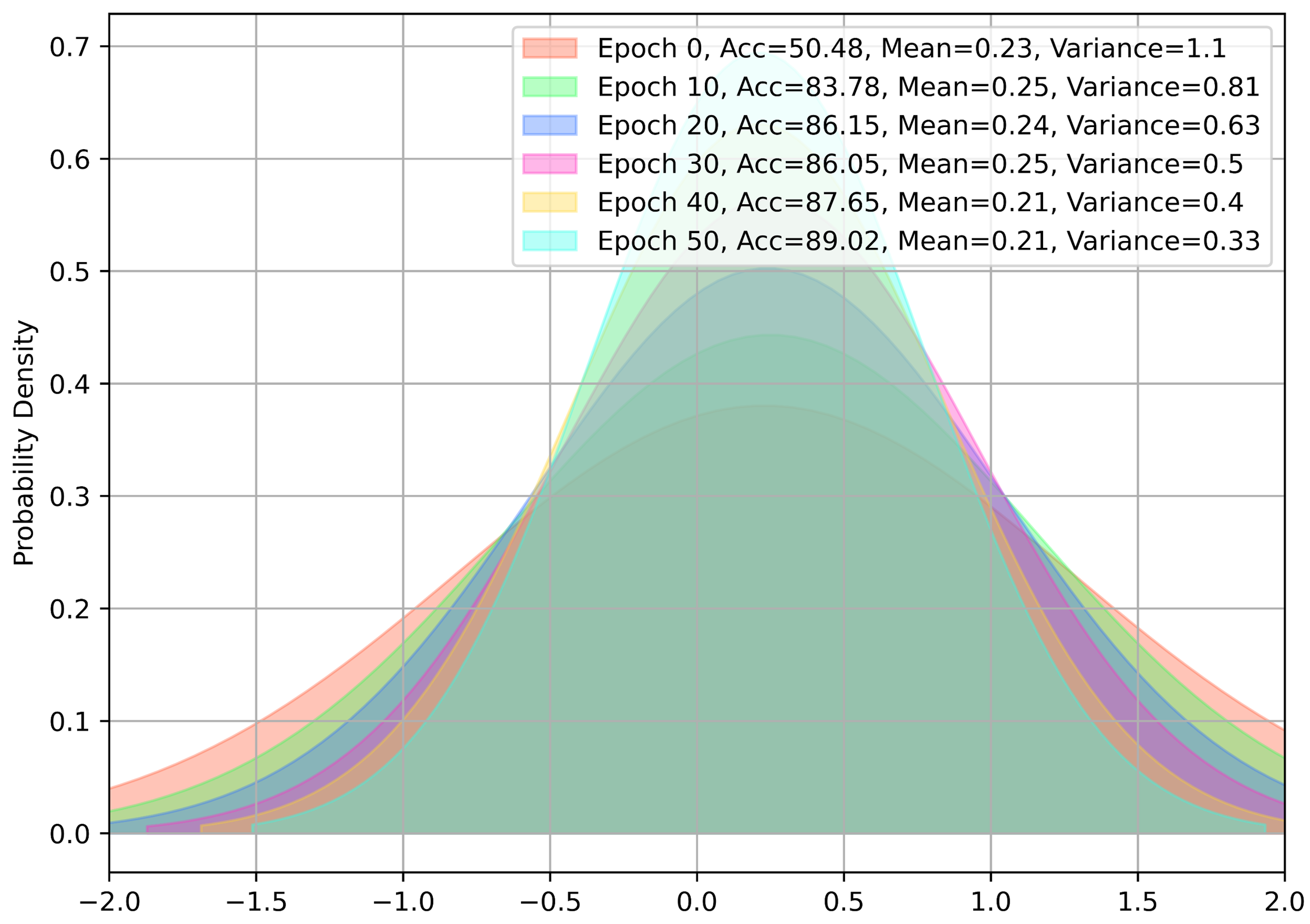}}
    \hspace{0.02\textwidth}
    \subfloat[]{\includegraphics[width=0.45\textwidth]{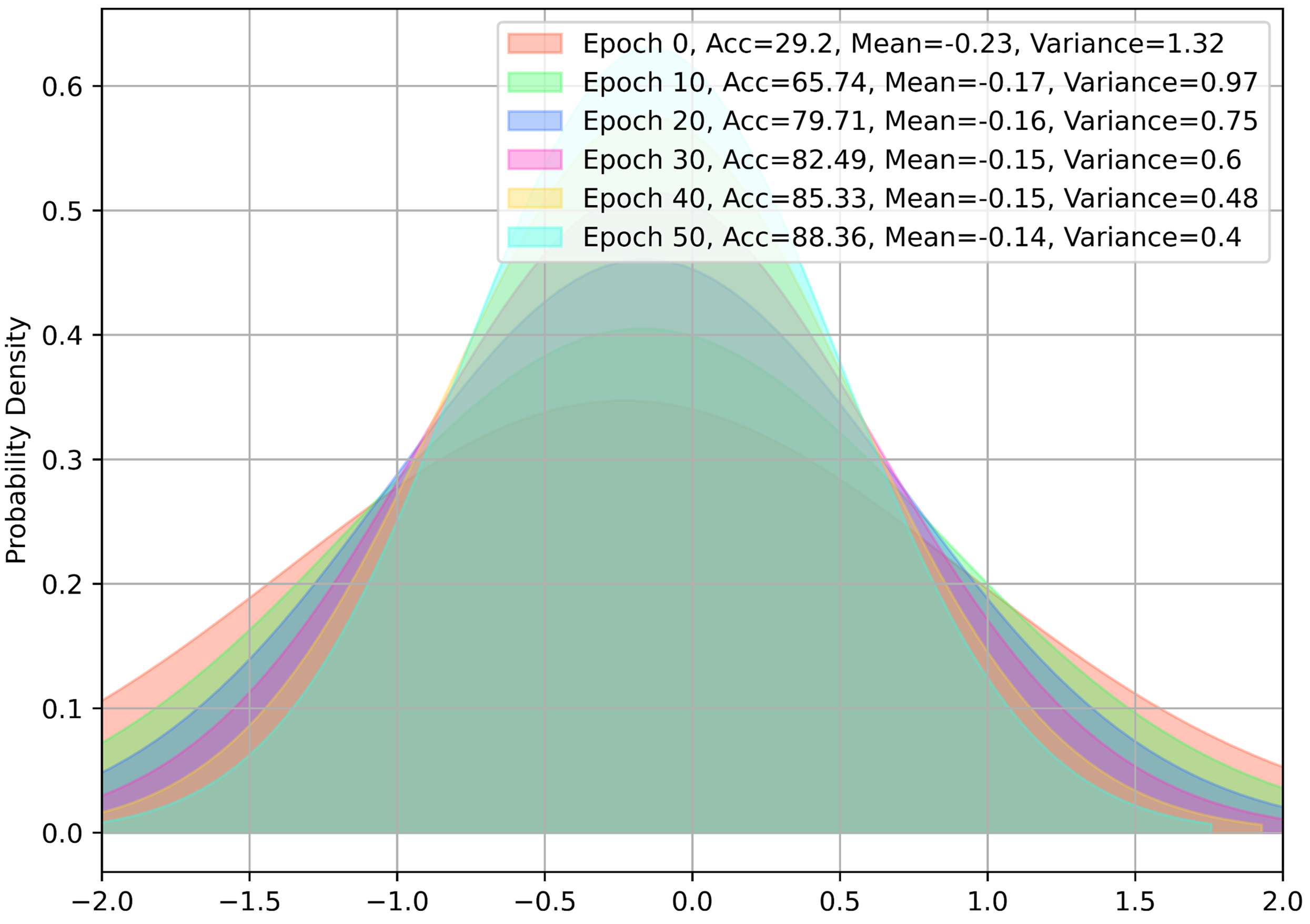}}
    \caption{Probability density estimates of the mean and variance of the global latent variable and the accuracy of testing on the target domain. (a) Source domain C0 is transferred to target domain C1. (b) Source domain C0 is transferred to target domain C2. More transfer results are provided in \textbf{Figure 8} of the Appendix.}
    \label{fig_5}
\end{figure}

\textit{\textbf{Analysis of global uncertainties at model level}}: The global uncertainty refers to a measure that indicates the average output error of the model in the entire dataset, which can utilize the statistics of the global latent variable to capture. Through the change of the statistics during the training, the modification of model uncertainty can be obtained. To further illustrate the global uncertainty, we show the changes in the probability density curve of the global latent variables in Figure \ref{fig_5}. Each subplot corresponds to a distinct experiment. A subplot includes information about the epoch, the mean and the variance of the global latent variable. The test accuracy of the GTNP in the target domain is also supplied in each subplot. Different curves in a subplot represent the probability density curves of global latent variables at different epochs in an experiment. As the epoch increases, the curve exhibits minimal shifting. It gradually transitions from a state of greater breadth to one of increased slimness, indicating that the mean remains relatively stable while the variance steadily decreases. Meanwhile, the test accuracy of the GTNP is enhanced.

The global uncertainty is influenced by the representation ability of the model and the complexity of the task \cite{wang2022uncertainty}, which can be further verified by experiments in the aircraft dataset. The global latent variable represents the average measure of the designed model. Specifically, as the epoch increases, the model fits the data better than initially, and the output error decreases consequently, resulting in the variance decreasing steadily. In general, a model with a larger parameter scale can fit more complex data distributions and obtain higher accuracy. However, a larger parameter scale introduces more accumulation of output error and results in higher variance. In order to demonstrate the above conclusion, we conducted 12 transfer tasks employing 2 models characterized by different scale parameters (additional information regarding the 2 models can be found in Appendix \textbf{Table 7} and \textbf{Table 8}). We recorded the alterations in the test accuracy, mean and variance of the global latent variables. The results are shown in Table \ref{tab:5}. Notably, we highlight variance with bold fonts, signifying instances of higher variance within identical tasks. The models with larger scale parameters generally have larger variance and also achieve higher test accuracy. In summary, the analysis of the global uncertainty provides an overall understanding of the model and the tasks.

\begin{table}[!htbp]
\centering
\caption{Global uncertainties at model level}
\begin{tabular}{ccccccc}
\toprule
\multirow{2}{*}{Task} & \multicolumn{3}{c}{Lighter model} & \multicolumn{3}{c}{Model used in this paper} \\
\cmidrule(lr){2-4} \cmidrule(lr){5-7}
 & Mean & Variance & Accuracy & Mean & Variance & Accuracy \\
\midrule
C0$\rightarrow$C1 & 0.01 & 0.24 & 87.07 & 0.21 & \textbf{0.33} & 89.02 \\
C0$\rightarrow$C2 & -0.04 & 0.35 & 83.69 & -0.14 & \textbf{0.4} & 88.36 \\
C0$\rightarrow$C3 & 0.02 & \textbf{0.35} & 87.29 & -0.02 & 0.34 & 90.24 \\
C1$\rightarrow$C0 & 0.13 & 0.44 & 84.48 & 0.43 & \textbf{0.45} & 86.95 \\
C1$\rightarrow$C2 & -0.05 & 0.29 & 89.19 & -0.23 & \textbf{0.33} & 93.45 \\
C1$\rightarrow$C3 & -0.08 & 0.33 & 87.36 & -0.25 & \textbf{0.34} & 91.77 \\
C2$\rightarrow$C0 & 0.04 & 0.29 & 85.45 & -0.54 & \textbf{0.45} & 87.13 \\
C2$\rightarrow$C1 & -0.004 & \textbf{0.37} & 85.55 & -0.23 & 0.33 & 91.01 \\
C2$\rightarrow$C3 & 0.32 & 0.27 & 90.97 & -0.31 & \textbf{0.36} & 93.32 \\
C3$\rightarrow$C0 & -0.04 & 0.26 & 85.2 & 0.45 & \textbf{0.41} & 87.13 \\
C3$\rightarrow$C1 & 0.27 & 0.27 & 86.28 & 0.19 & \textbf{0.4} & 91.75 \\
C3$\rightarrow$C2 & 0.11 & 0.37 & 91.1 & -0.21 & \textbf{0.39} & 93.91 \\
\midrule
Average & 0.06 & 0.32 & 86.97 & -0.05 & \textbf{0.38} & 90.34 \\
Standard Deviation & 0.12 & \textbf{0.06} & 2.29 & 0.3 & 0.04 & 2.48 \\
\bottomrule
\end{tabular}
\label{tab:5}
\end{table}

\textit{\textbf{Analysis of local uncertainties at sample level}}: Local uncertainty refers to a measure that evaluates the confidence of prediction through the local latent variable for each sample. Unlike utilizing the statistics of the global latent variable to capture the global uncertainty, we employ the Monte Carlo Random Sampling method \cite{shapiro2003monte} to resample 100 times from $p(u|x)$ since the local latent variable is only utilized in the forward inference prediction. The final result is predicted by the normalized value derived of the log probability from the 100 iterations for the test sample.

\begin{figure*}[!h]
\centering
\includegraphics[width=5.5in]{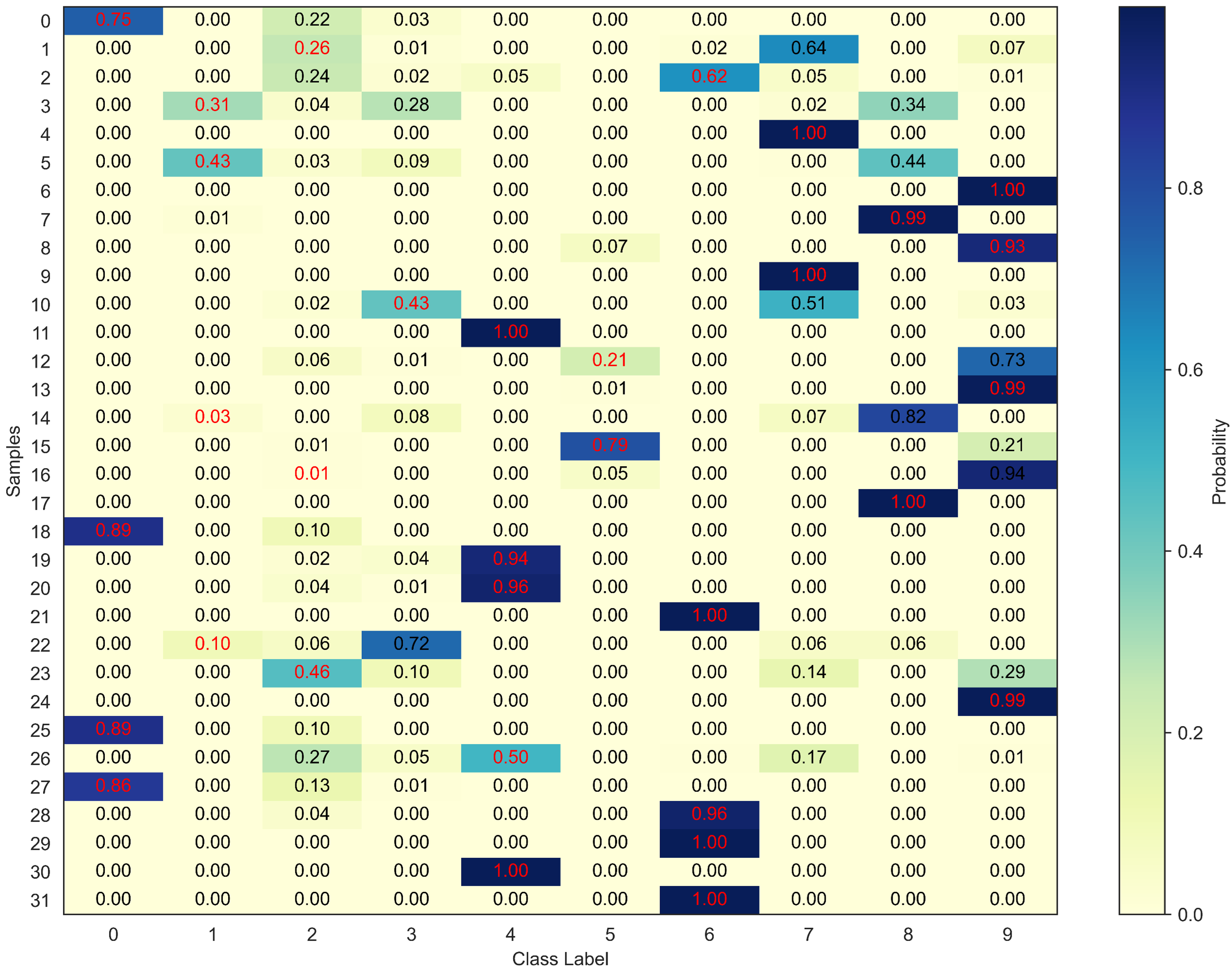}
\caption{Local uncertainties based on Monte Carlo Random Sampling (the number of resampling is set to 100).}
\label{fig_6}
\end{figure*}

\begin{figure}[!h]
    \centering
    \subfloat[]{\includegraphics[width=0.45\textwidth]{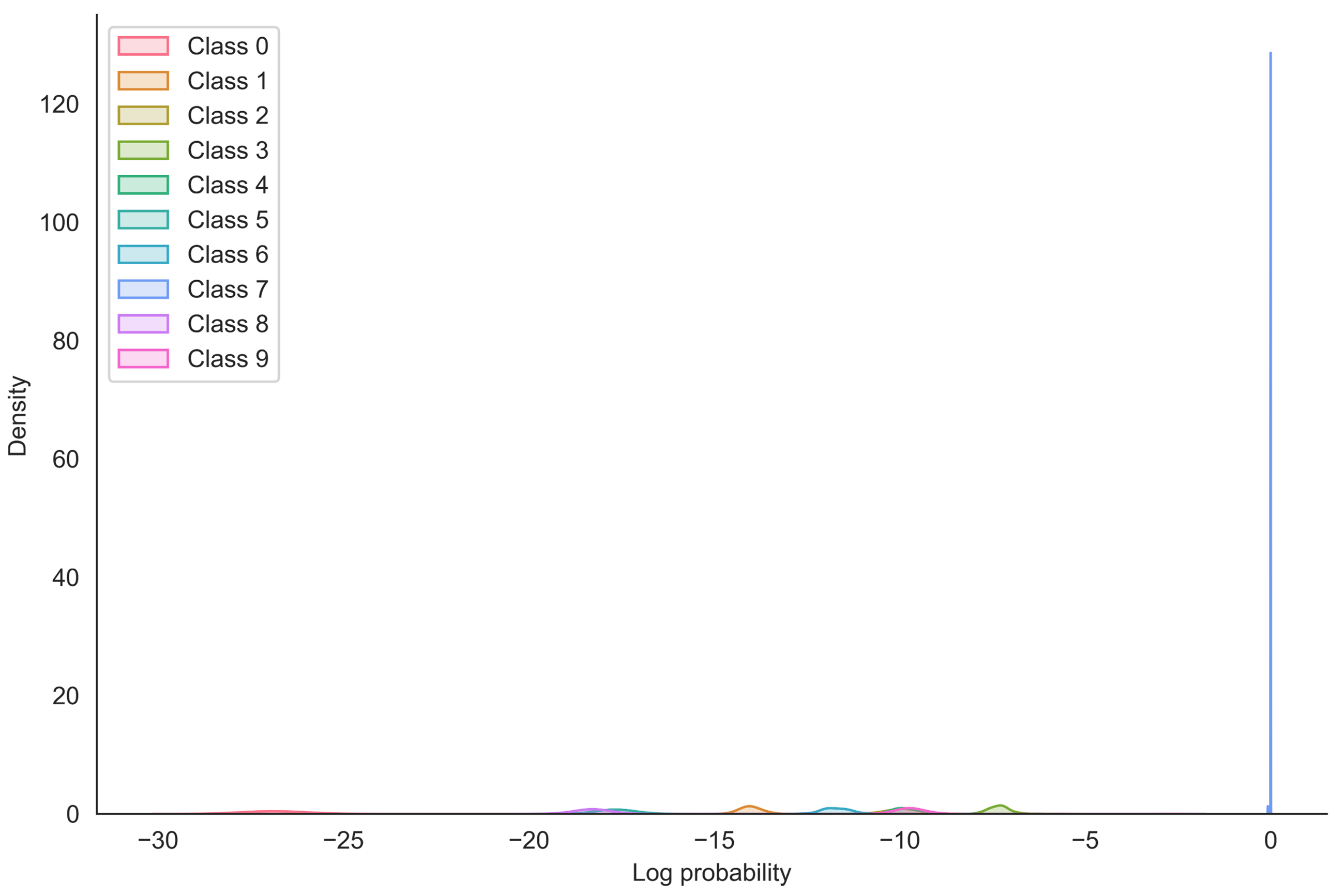}
    \label{fig_7a}
    }
    \hspace{0.02\textwidth}
    \subfloat[]{\includegraphics[width=0.45\textwidth]{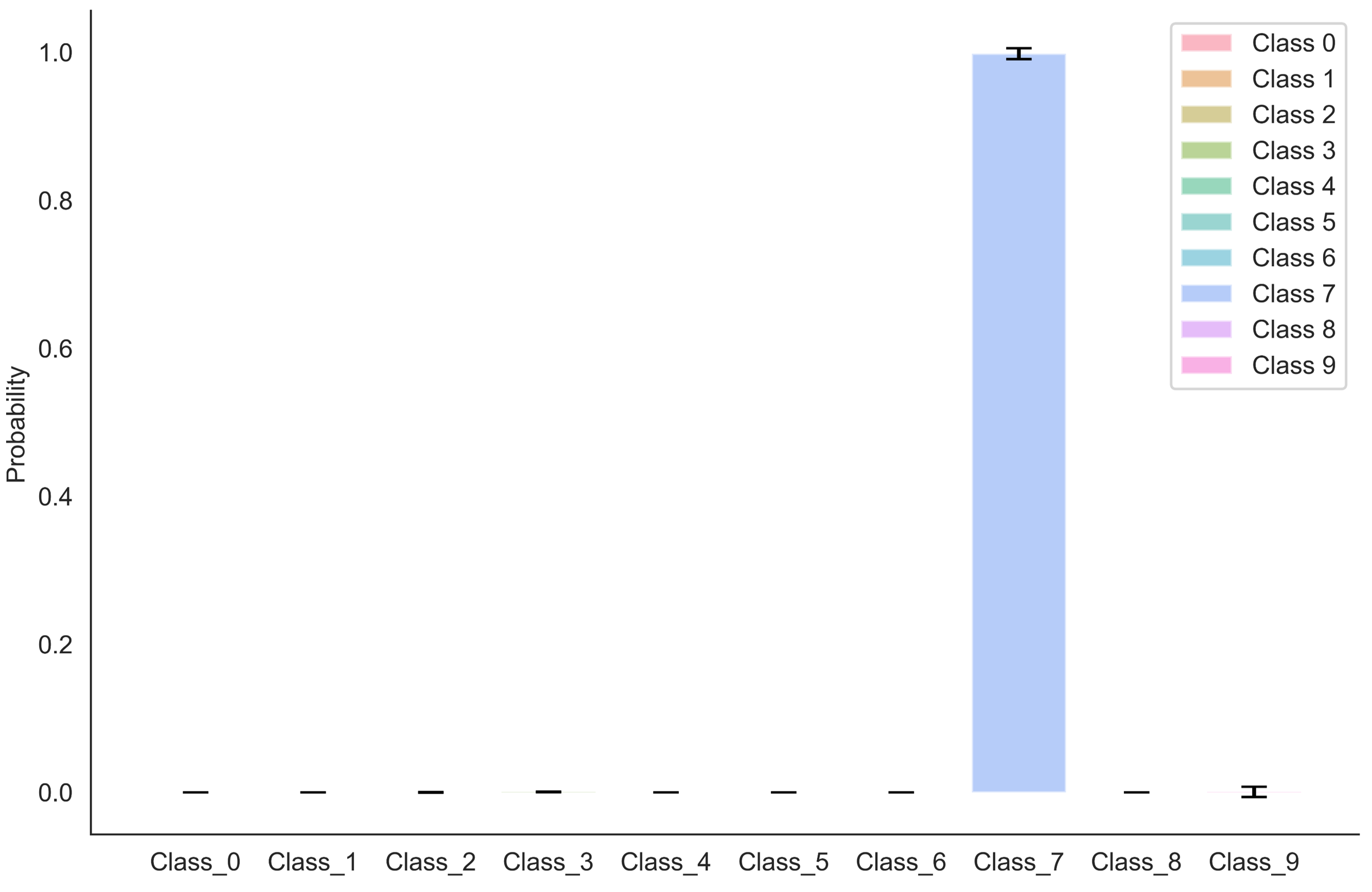}
    \label{fig_7b}
    }\\
    \subfloat[]{\includegraphics[width=0.45\textwidth]{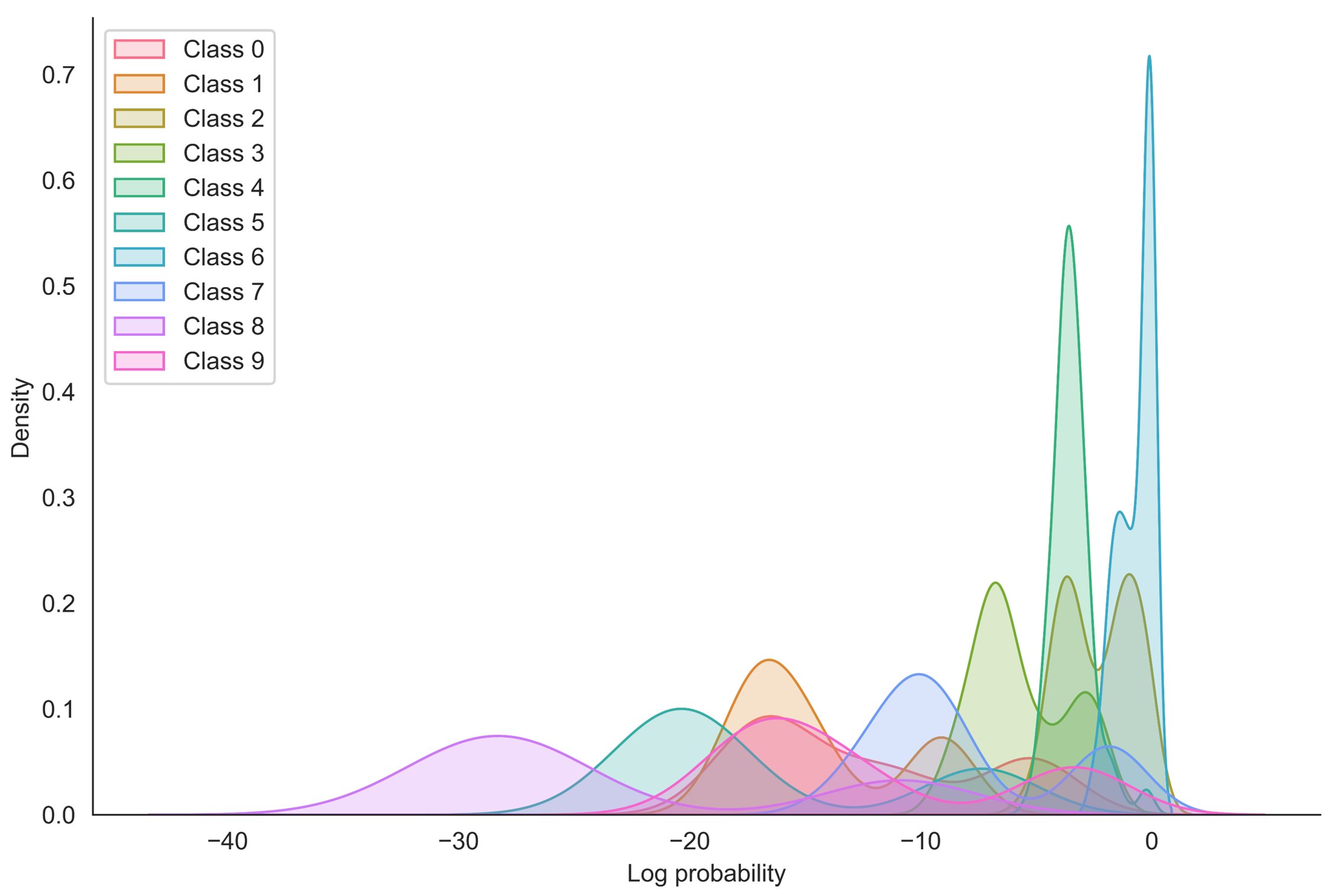}
    \label{fig_7c}
    }
    \hspace{0.02\textwidth}
    \subfloat[]{\includegraphics[width=0.45\textwidth]{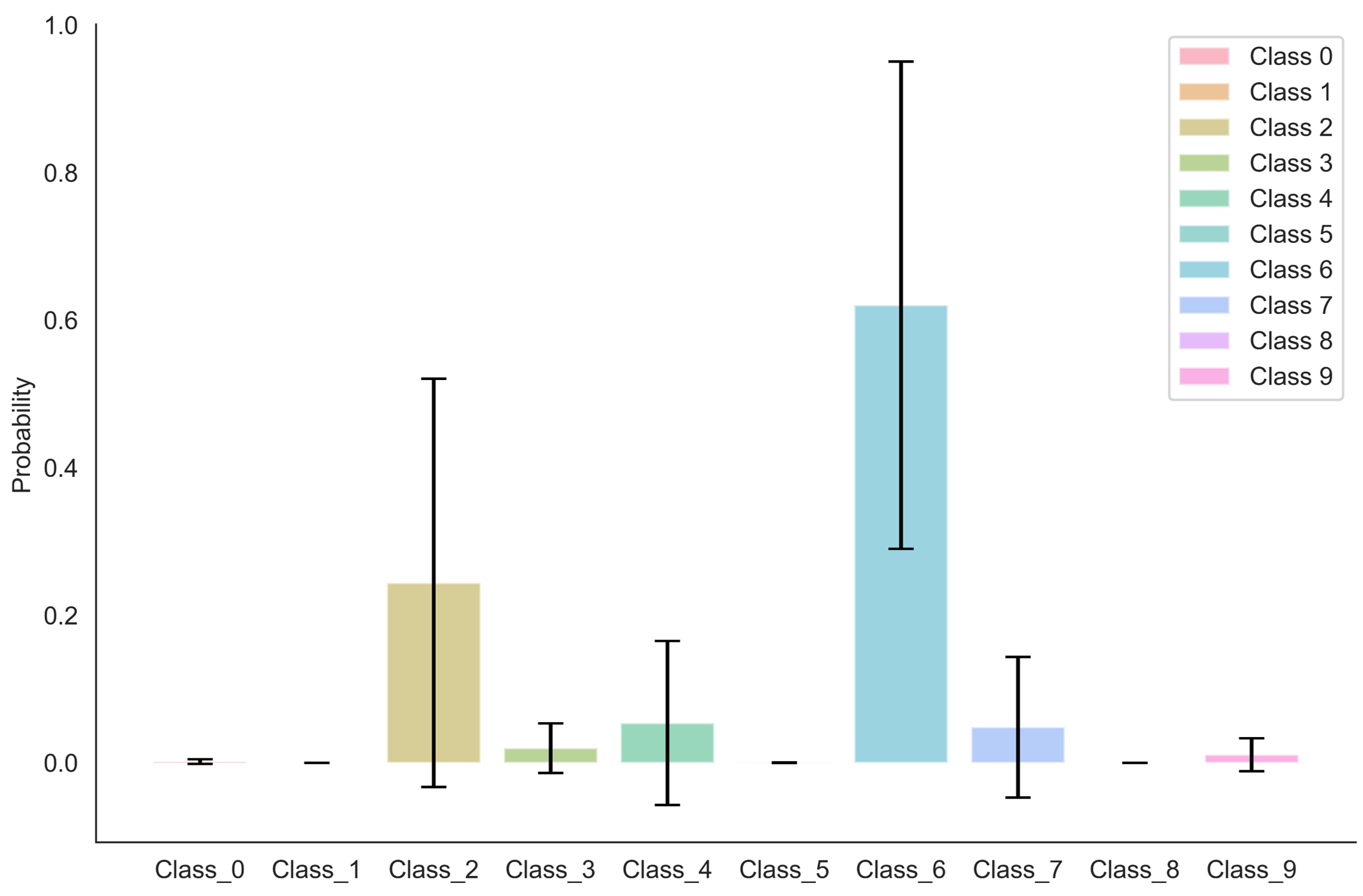}
    \label{fig_7d}
    }\\
    \subfloat[]{\includegraphics[width=0.45\textwidth]{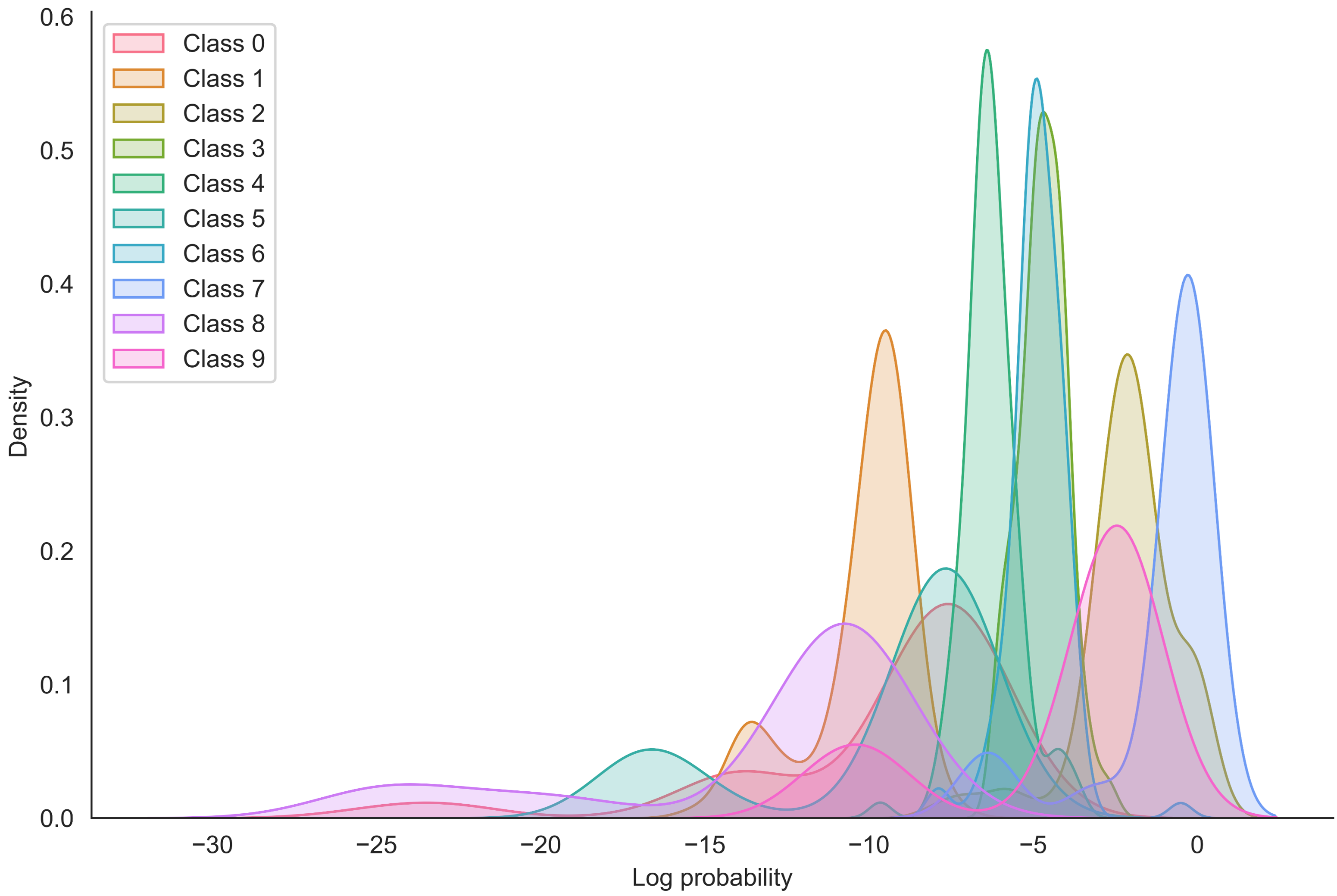}
    \label{fig_7e}
    }
    \hspace{0.02\textwidth}
    \subfloat[]{\includegraphics[width=0.45\textwidth]{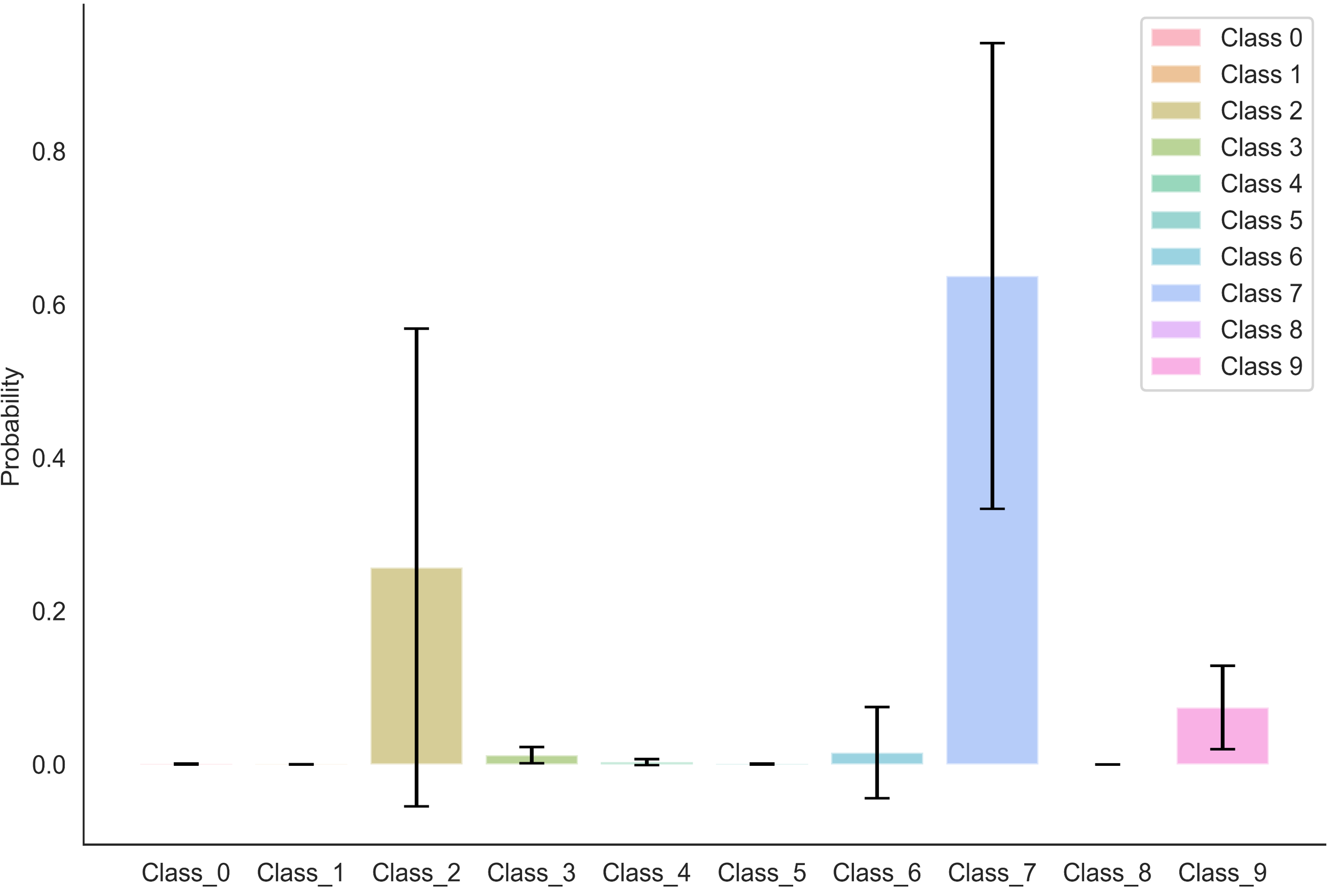}
    \label{fig_7f}
    }
    \caption{Local uncertainties at sample level are determined using the Monte Carlo Random Sampling method (with consistent colors for the same class in the two subplots of the same row). Subplots (a), (c), and (e) present probability density estimation results for each class, while subplots (b), (d), and (f) display the results of normalizing of log probability statistics.}
    \label{fig_7}
\end{figure}

The predicted probability of 32 test samples under each class in the $20^{th}$ iteration are shown in Figure \ref{fig_6}. The greater the prediction probability, the darker the color of the square, and the red font corresponds to the real class according to the label. Though traditional neural networks can also obtain similar results, GTNP is based on distribution which provides extra support for local uncertainty analysis at the sample level. The detailed results are shown in Figure \ref{fig_7}.

To further illustrate the local uncertainty clearly, Figure \ref{fig_7} visualizes the classification probabilities derived from 100 random samplings, corresponding to sample indexes 9, 2, and 1 in Figure \ref{fig_6}. Specifically, Figures \ref{fig_7a}, \ref{fig_7c}, and \ref{fig_7e} display the results of probability density estimation for each class, while Figures \ref{fig_7b}, \ref{fig_7d}, and \ref{fig_7f} display the results of normalizing of log probability statistics. In Figures \ref{fig_7a} and \ref{fig_7b}, the sample is confidently classified as Class 7, which aligns with the label, indicating a low level of uncertainty in the prediction for this sample. However, the sample in Figures \ref{fig_7c} and \ref{fig_7d} is classified as Class 6 with a few degrees of uncertainty, which may impact the decision-making process. Particularly, the sample in Figures \ref{fig_7e} and \ref{fig_7f} is incorrectly classified as Class 7, in contrast to its true label of Class 2. Figures \ref{fig_7e} and \ref{fig_7f} reveal that the probability values for the sample under Class 7 exceed 0.8, while the predicted probabilities for Class 2 and Class 7 exhibit significant variance. The results are contributed by multiple factors, including noise in the dataset, leading to the GTNP overlapping relationships between the sample distributions of Class 2 and Class 7. As a solution, additional discriminative features should be designed to enhance the differentiation of samples between Class 2 and Class 7. In summary, the GTNP provides local uncertainty based on the distribution perspective and gives an extra local uncertainty understanding compared to the traditional models.

\subsection{Case study of sensor fault detection with different aircraft}
\subsubsection{Construct the representation of knowledge in source domain}
\begin{figure*}[!h] 
    \centering
    \begin{subfigure}[b]{0.42\textwidth}
        \includegraphics[width=\textwidth]{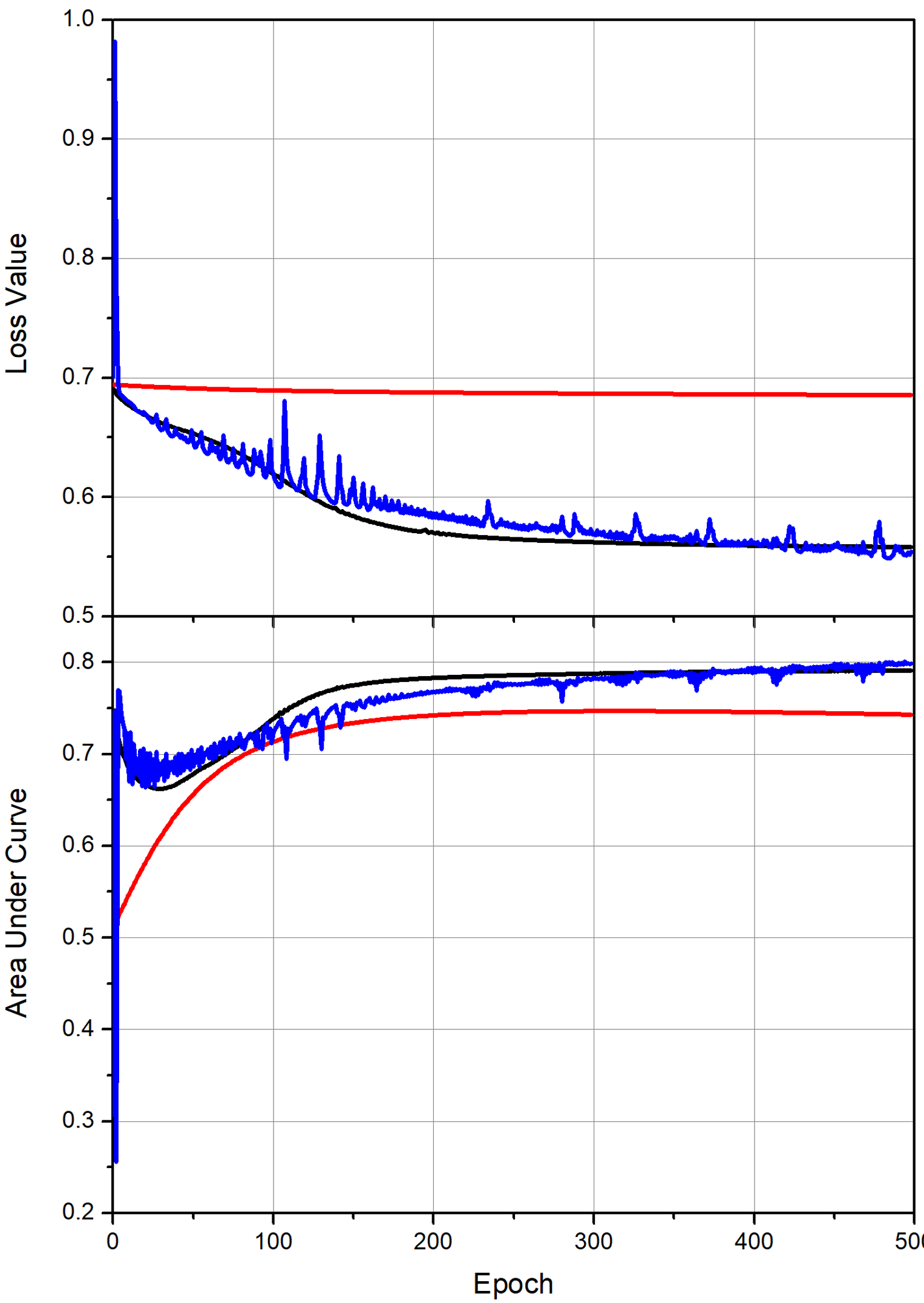}
        \caption{}
    \end{subfigure}
    \begin{subfigure}[b]{0.528\textwidth}
        \includegraphics[width=\textwidth]{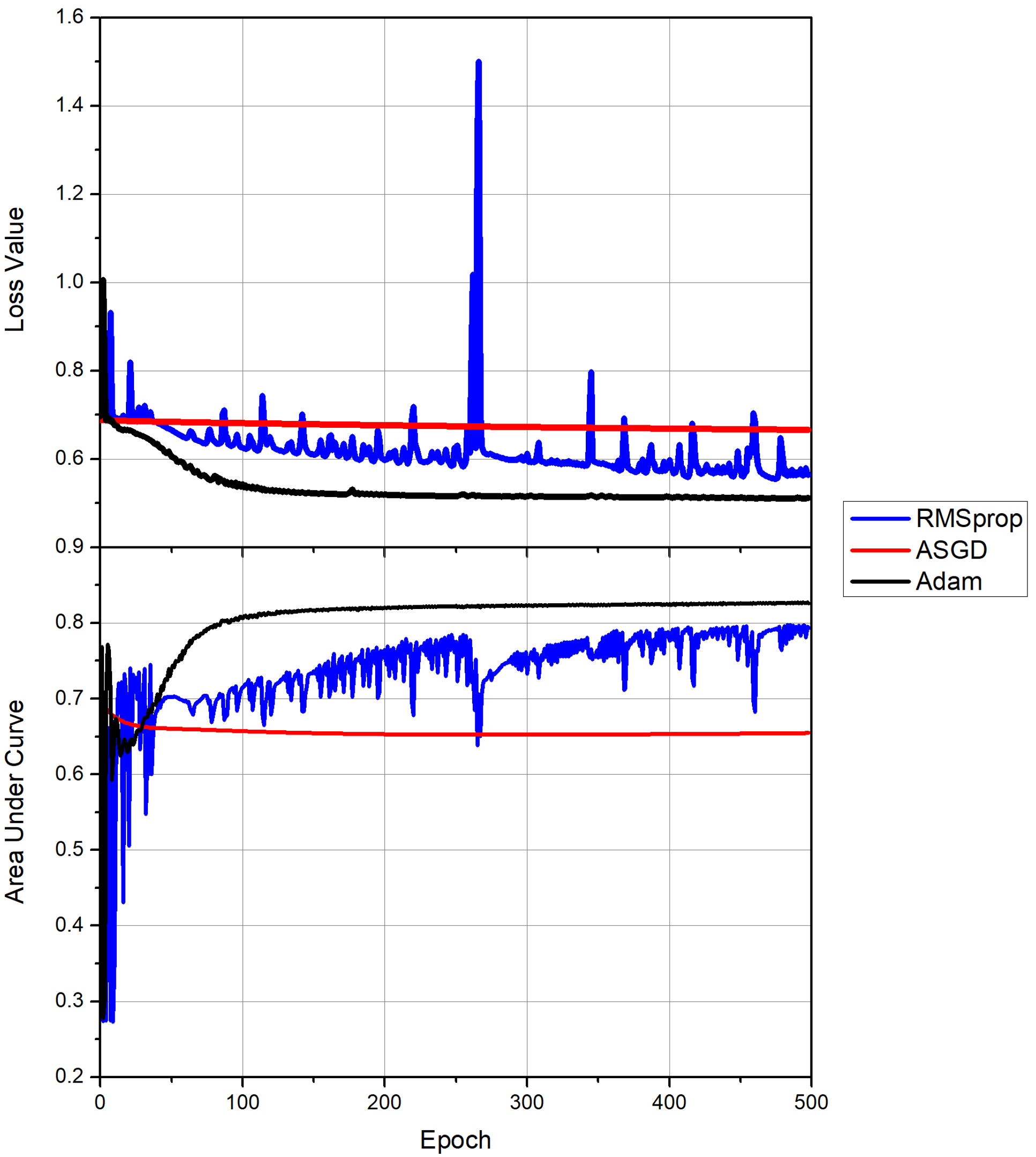}
        \caption{}
    \end{subfigure}
    \caption{GCN training results on aircraft dataset. (a) The learning rate is set to 0.001. (b) The learning rate is set to 0.01.}
    \label{fig_8}
\end{figure*}

For the aircraft sensor fault detection dataset, which comprises data from 3 simulating aircraft (B1, D, Y) and 1 real aircraft (F), we employ a similar approach. Specifically, GCN is employed to construct detection knowledge in the source domain. We randomly select 300 samples from the source domain (B1 dataset) as the training set for the GCN. The impact of the learning rate and the optimizer of the GCN is shown in Figure \ref{fig_8}.

As a consequence of the aircraft dataset's low signal-to-noise ratio, the training results do not attain the same level as those on the CWRU dataset. However, some commonalities can be found between the 2 datasets. For the learning rates, a learning rate of 0.001 impedes the optimization process, leading to adverse effects on accuracy convergence. Furthermore, the ADAM optimizer proves to be the most effective choice for optimizing the GTNP on the aircraft dataset. Consequently, the learning rate of 0.01 and the Adam optimizer are selected in the experiments for the aircraft dataset.

\subsubsection{Analysis of ablation comparison experiments}
\begin{figure*}[!h]
\centering
\includegraphics[width=4.5in]{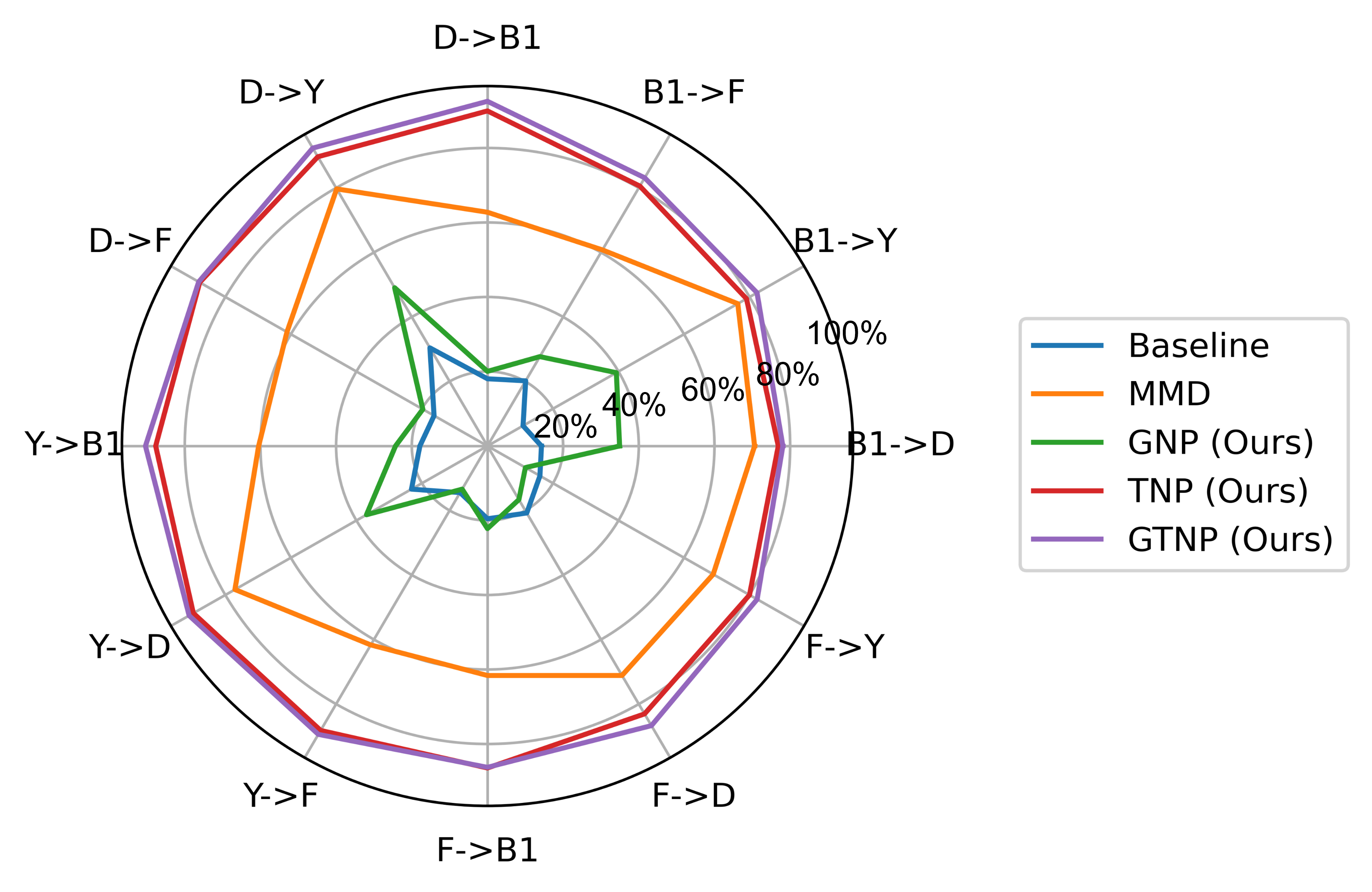}
\caption{Ablation comparison experimental results of 12 transfer tasks on the aircraft dataset.}
\label{fig_9}
\end{figure*}

To validate the effectiveness of the GTNP proposed in this paper, we organize a series of ablation comparison experiments in this section. For the source domain, we maintain an 8:2 ratio to divide the training and testing sets. Additionally, only 600 samples are randomly selected from the target domain for the training set (comprising 2.3\%, 2.2\%, 2.1\%, and 6.5\% of B1, D, Y, and F, respectively), and the remaining samples construct the testing set (comprising 97.7\%, 97.8\%, 97.9\%, and 93.5\% of B1, D, Y, and F, respectively). The methods employed in the ablation comparison experiments include the followings: (1) a trained baseline network (ResNet-18) only utilizing data in source domain, (2) an MMD based DTL method, (3) the graph neural process (GNP) without the incorporation of DTL method, and (4) the transfer neural process (TNP) without GCN. For a fair comparison, all methods in this section are configured with the same parameters, including batch size, optimizer, and learning rate. Each experiment is repeated 10 times, and the average results for each method are shown in Figure \ref{fig_9} and Table \ref{tab:4}.

\begin{table}[!ht]
\centering
\caption{The average accuracy ($\%$) on the aircraft dataset with different methods.}
\label{tab:6}
\begin{adjustbox}{width=\textwidth} 
\begin{tabular}{cccccccc}
\hline
Task & ResNet-18 & MMD & GNP (Ours) & TNP (Ours) & GTNP (Ours) \\ \hline
B1$\rightarrow$D & 14.27 & 70.61 & 34.92 & 76.87 & \textbf{78.07} \\ \hline
B1$\rightarrow$Y & 10.86 & 76.4 & 39.35 & 79.06 & \textbf{82.27} \\ \hline
B1$\rightarrow$F & 20.2 & 60.78 & 27.7 & 80.56 & \textbf{83.13} \\ \hline
D$\rightarrow$B1 & 18.04 & 62.75 & 20.03 & 89.95 & \textbf{92.53} \\ \hline
D$\rightarrow$Y & 30.42 & 79.69 & 49.06 & 89.65 & \textbf{92.34} \\ \hline
D$\rightarrow$F & 16.29 & 61.19 & 19.72 & 87.82 & \textbf{88.12} \\ \hline
Y$\rightarrow$B1 & 17.86 & 60.47 & 24.3 & 87.69 & \textbf{90.39} \\ \hline
Y$\rightarrow$D & 23.19 & 77.06 & 36.92 & 89.76 & \textbf{91.04} \\ \hline
Y$\rightarrow$F & 14.45 & 61.67 & 13.38 & 88.17 & \textbf{89.39} \\ \hline
F$\rightarrow$B1 & 19.54 & 61.61 & 22.18 & 86.44 & \textbf{86.26} \\ \hline
F$\rightarrow$D & 20.77 & 71.13 & 16.62 & 83.04 & \textbf{86.68} \\ \hline
F$\rightarrow$Y & 16.02 & 68.9 & 11.59 & 79.89 & \textbf{82.31} \\ \hline
Average & 18.49 & 67.69 & 26.31 & 84.92 & \textbf{86.54} \\ \hline
\end{tabular}
\end{adjustbox}
\end{table}

\begin{figure}[!t]
\centering
\includegraphics[width=6in]{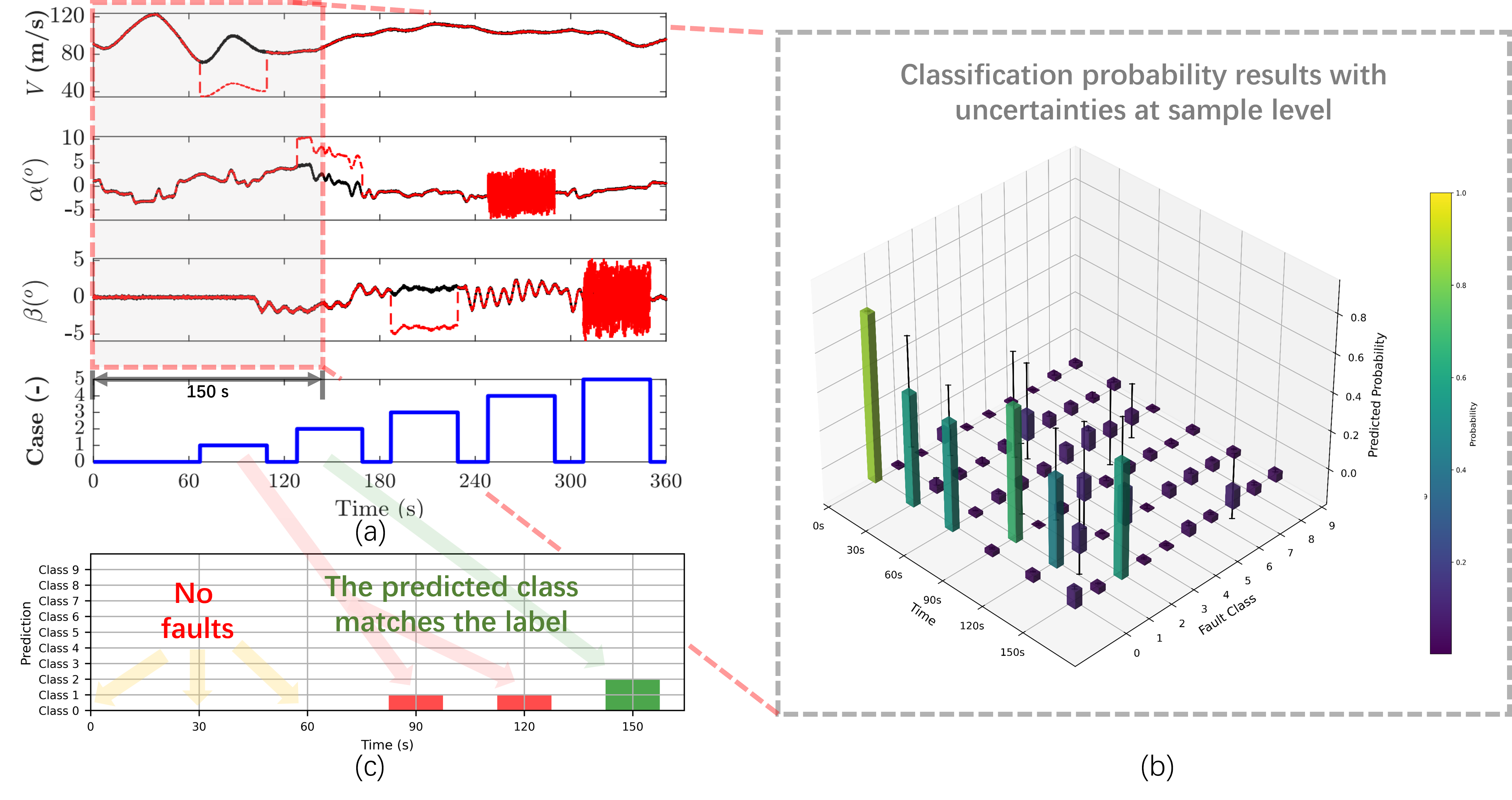}
\caption{The application of proposed GTNP for IFD tasks. (a) The clean states, originating from real or simulated aircraft, are depicted in black, while the red lines represent data with intentionally injected faults. The primary objective of the IFD tasks are to effectively identify the various fault cases (indicated by the bold blue line). (b) The classification probability results of data spanned a time duration of 150$s$ at different time points (0$s$, 30$s$, 60$s$, 90$s$, 120$s$, and 150$s$). (c) The final classification results output by GTNP.}
\label{fig_A3}
\end{figure}

In this section, we focus on evaluating the effectiveness of GTNP in detecting sensor faults between various aircraft on the aircraft dataset, thereby assessing its generalization ability. The average accuracy of GTNP across 12 transfer experiments attains an impressive 86.54\%, marking a significant improvement of 68.05\% compared to the baseline method. Notably, due to the variations of data distribution between distinct aircraft, the baseline algorithm in this section achieves an average classification accuracy of 18.49\%, which is considerably lower compared to the results obtained in the CWRU dataset (as depicted in Table \ref{tab:4}). Ablation experiments carried out among GNP, TNP, and GTNP elucidate their comparison performance. GNP, despite its absence of transfer learning, shows some enhancement compared to the baseline method but remains unsuitable for practical IFD tasks. TNP, which overlooks the incorporation of source domain knowledge, exhibits a 1.62\% lower average accuracy compared to the proposed GTNP. This validates the effectiveness of utilizing GCN for enhancing detection knowledge representation in the source domain and subsequently transferring it to the target domain. The radar chart illustrated in Figure \ref{fig_9} reveals that MMD methods exhibit low accuracy in transfer tasks, such as B1 to F, D to F, and Y to F (in comparison to other transfer tasks such as B1 to Y and B1 to D). This disparity is attributed to the fact that F, as the real aircraft data, is characterized by a smaller dataset size, higher noise, and more complex data distribution than the data of other simulating aircraft. This problem is effectively mitigated by TNP and GTNP that proposed in this paper. The role of the reference set and the multi-scale modeling strategy not only reduces the demand number of data in the target domain but also enhances the representational characteristics of samples in both source and target domains. Figure \ref{fig_A3} exemplifies the utilization of the GTNP model in a practical application. In this context, the GTNP model is employed to compute probability values for measurement data containing 2 faults. The resultant output corresponds to the class with the highest probability value, which is consistent with the real label. In summary, the method proposed in this paper demonstrates significant performance on the aircraft dataset, highlighting its superior capability in handling transfer tasks across different aircraft when compared to other ablation comparison methods.

\subsubsection{Analysis of global uncertainties at model level}
\begin{figure*}[!h] 
    \centering
    \begin{subfigure}[b]{0.48\textwidth}
        \includegraphics[width=\textwidth]{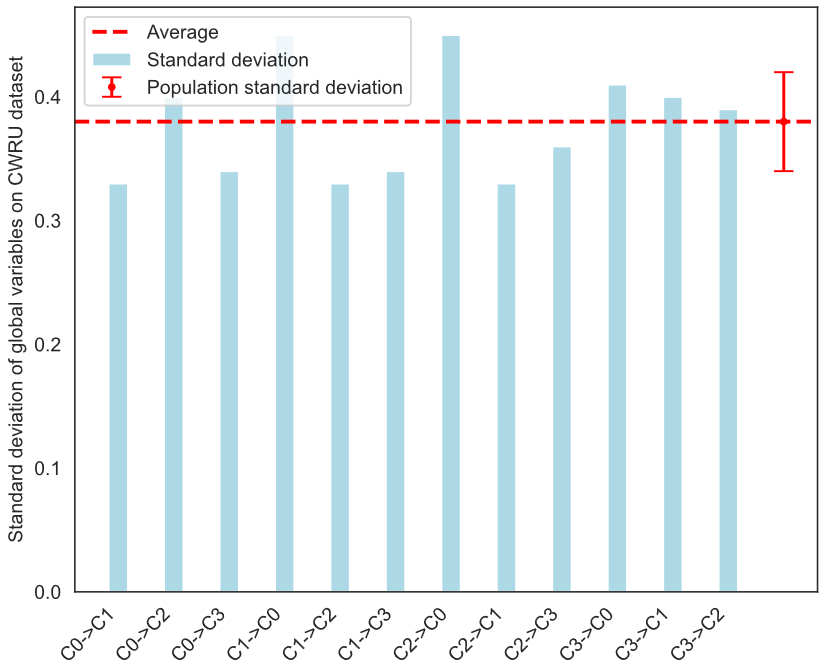}
        \caption{}
        \label{fig_10a}
    \end{subfigure}
    \begin{subfigure}[b]{0.48\textwidth}
        \includegraphics[width=\textwidth]{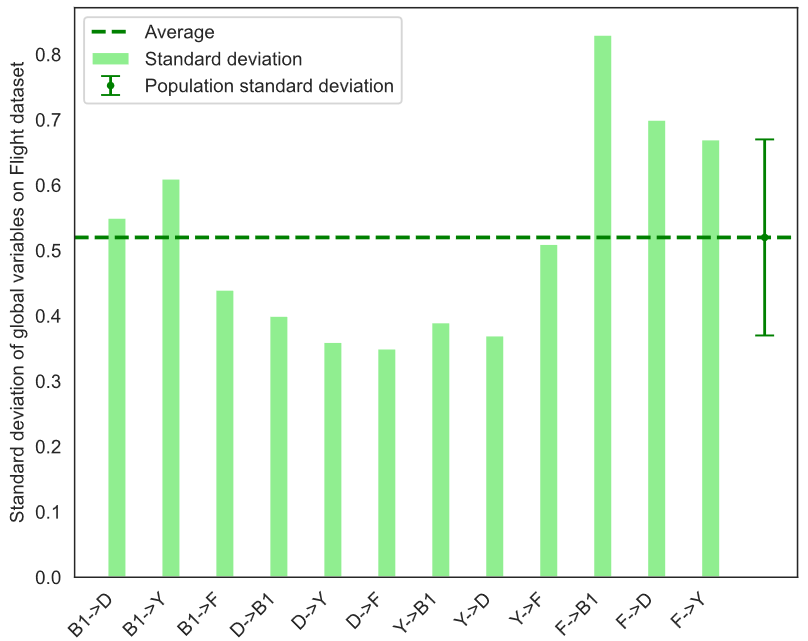}
        \caption{}
        \label{fig_10b}
    \end{subfigure}
    \caption{The results of global latent variable variances between different tasks. (a) Global latent variable variances and statistical results under 12 transfer tasks on the CWRU dataset. (b) Global latent variable variances and statistical results under 12 transfer tasks on the aircraft dataset.}
    \label{fig_10}
\end{figure*}

As delineated in Section 2.5.5, the variance of global latent variables, which serves to capture the global uncertainty, exhibits a discernible relationship with the difficulty of the transfer task. We conduct a total of 12 transfer experiments on the CWRU dataset and observe an average variance of 0.38 and a standard deviation of 0.04. However, as shown in Figure \ref{fig_10}, for the 12 transfer experiments on the aircraft dataset, the average global latent variable variance is 0.52 and the standard deviation is 0.15. The results further validate the prior conclusion that transfer experiments involving different aircraft pose a greater challenge. Notably, Figure \ref{fig_10b} shows that the variance of global latent variables during transfer experiments between 3 simulating aircraft is lower than the variance during transfer tasks from a real aircraft to 3 simulating aircraft. The result demonstrates that the more complex tasks yield global latent variables with higher variances. Therefore, the global variables are used to assess both the complexity of model and the inherent difficulty of transfer tasks.

\subsection{Case study of emerging sensor fault detection}
During the service of machinery, it is common for new and unforeseen faults to occur, especially for machines such as aircraft that operate in harsh environments. Therefore, it is essential to investigate the application of deep transfer learning for the detection of new faults in the aircraft dataset. The dataset presented in Table \ref{tab:7} corresponds to the data collected during a series of real aircraft operations. The numbers 1-9 represent fault types present in the source domain (corresponding to the fault types in B1, D, Y, and F), while the number 0 indicates a new sensor fault (specifically, an angle of attack sensor faults). In this section, we employ the F dataset as the source domain (both F and the new data originate from real aircraft). We utilize the newly collected dataset (Table \ref{tab:7}) as the target domain. Based on the approach discussed above, the training and testing sets are partitioned according to an 8:2 ratio in the source domain, and 600 samples are randomly selected from the new dataset of the target domain as the training set (300 samples in the reference set, maintaining consistency with the previous experiments), while all remaining samples are designated as testing set, constituting 87\% of the total samples. Detailed information can be found in Table \ref{tab:6}.

\begin{table}[!h]
\centering
\caption{The new dataset used for emerging sensor fault detection.}
\begin{tabular}{cccc}
\hline
\textbf{ } Fault type & Collected data & Training data & Testing data \\
\hline
0 & 873 & 114 & 759 \\
1 & 556 & 61 & 495 \\
2 & 194 & 25 & 169 \\
3 & 416 & 59 & 357 \\
4 & 408 & 49 & 359 \\
5 & 437 & 69 & 368 \\
6 & 195 & 28 & 167 \\
7 & 517 & 74 & 443 \\
8 & 528 & 73 & 455 \\
9 & 335 & 48 & 287 \\
\hline
Sum & 4459 & 600 & 3859 \\
\hline
\end{tabular}
\label{tab:7}
\end{table}

\begin{figure}[!ht]
    \centering
    \subfloat[epoch 0]{\includegraphics[width=0.45\textwidth]{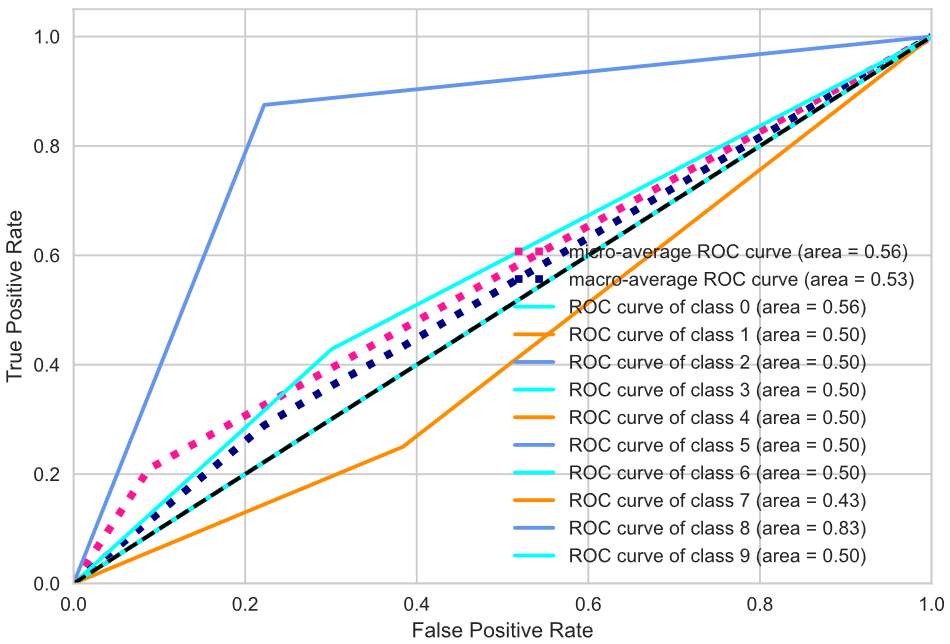}}
    \hspace{0.02\textwidth}
    \subfloat[epoch 10]{\includegraphics[width=0.45\textwidth]{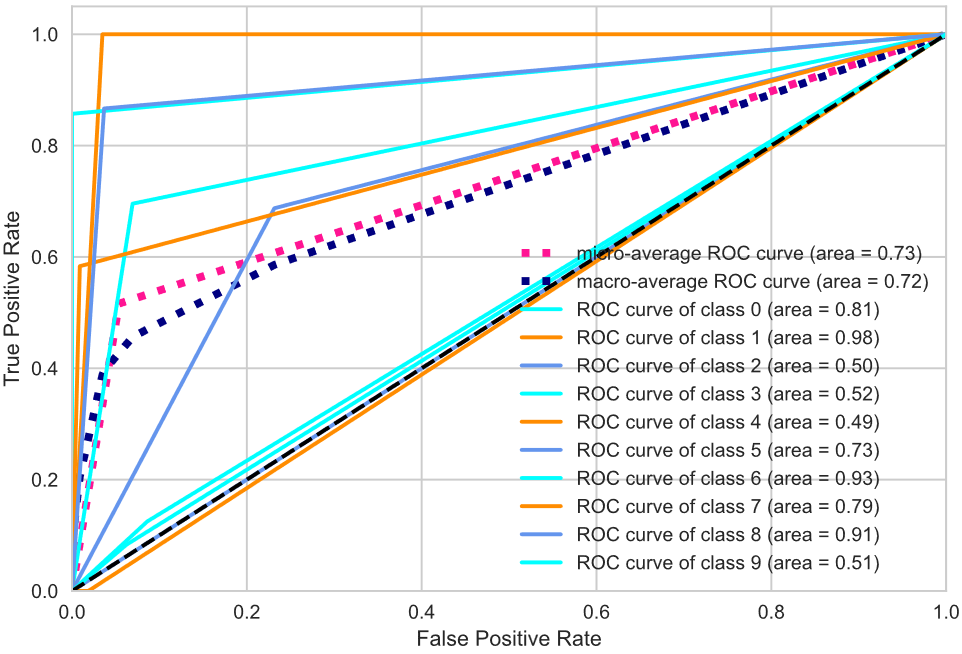}}\\
    \subfloat[epoch 20]{\includegraphics[width=0.45\textwidth]{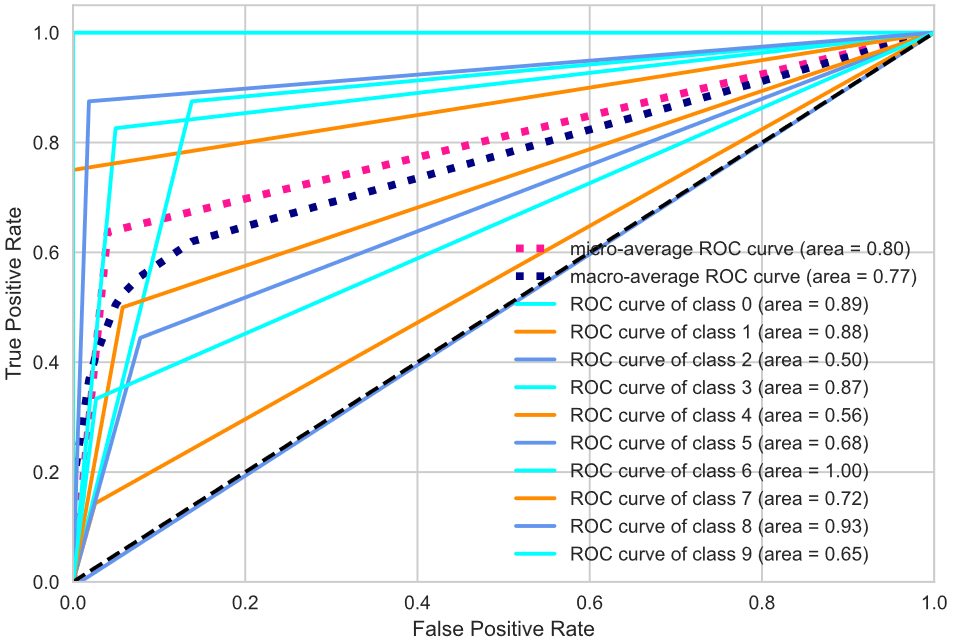}}
    \hspace{0.02\textwidth}
    \subfloat[epoch 30]{\includegraphics[width=0.45\textwidth]{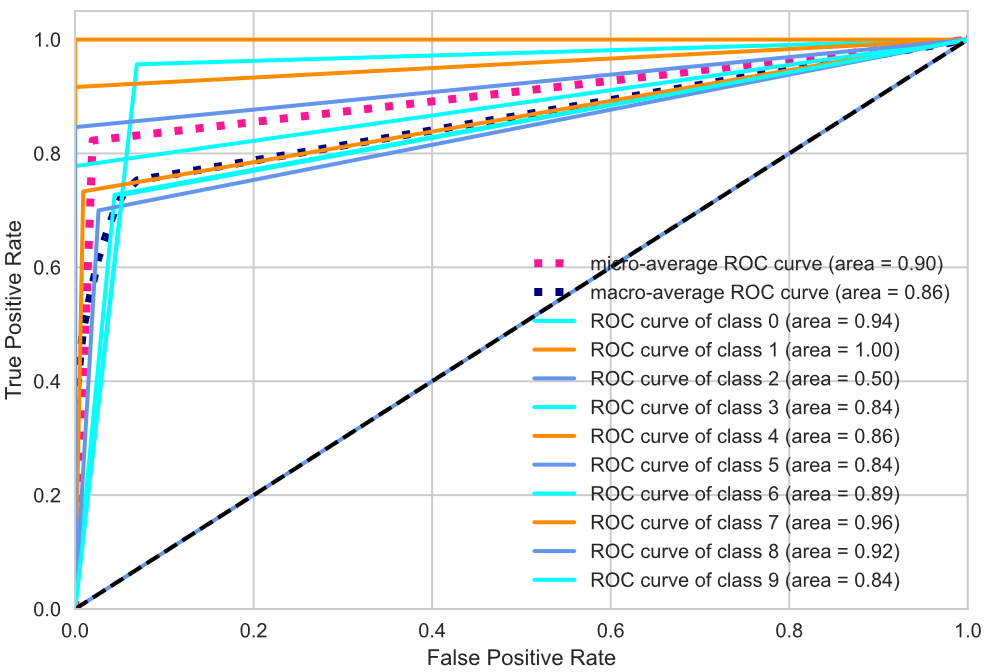}}
    \caption{ROC curve of GTNP model during testing on the target domain.}
    \label{fig_11}
\end{figure}

We select the roc curve to evaluate the test performance of the GTNP in the target domain under different epochs, which reflects the performance of the model in the case of class imbalance effectively since it takes TPR and FPR into account. Generally, $auc$ is used to quantify the overall performance of the roc curve. The larger the $auc$, the better the model performance. The test results are shown in Figure \ref{fig_11}. It can be found that as the iteration proceeds and the model parameters are optimized, the average classification performance of the model for all classes is improving. It is worth noting that the classification performance for class 0 (emerging sensor fault) is significantly improved (from 56\% to 94\%).

In order to comprehensively evaluate the performance of the method proposed in this paper, we designed a comparison experiment between GNP and GTNP, selecting $pre$, $rec$, and $f1$ as the evaluation metrics. We conducted 10 repeated experiments and calculated the average and standard deviation of each evaluation metric, which are as shown in Table \ref{table_7} and Table \ref{table_8}. It can be found that due to the prominent imbalance problem between different classes, the performance of the model in each class varies greatly. Notably, the average detection $f1$ values for new faults of GTNP is 87\%, while GNP achieved only 65\%. Additionally,  GTNP also performs better for the detection performance of other classes, except for a slight drop in class 8. The result verifies that GTNP effectively solves the emerging sensor fault detection problem in the target domain for IFD.

\begin{table}[!htbp]
\centering
\caption{Emerging sensor fault detection with GNP.}
\begin{tabular}{ccccccc}
\hline
Fault type & $pre$ & $rec$ & $f1$ \\
\hline
0 & 0.59 $\pm$ 0.048 & 0.728 $\pm$ 0.115 & 0.65 $\pm$ 0.073 \\
1 & 0.804 $\pm$ 0.07 & 0.922 $\pm$ 0.045 & 0.856 $\pm$ 0.039 \\
2 & 0.408 $\pm$ 0.221 & 0.296 $\pm$ 0.092 & 0.328 $\pm$ 0.121 \\
3 & 0.516 $\pm$ 0.146 & 0.518 $\pm$ 0.115 & 0.502 $\pm$ 0.092 \\
4 & 1 $\pm$ 0 & 0.284 $\pm$ 0.075 & 0.436 $\pm$ 0.087 \\
5 & 0.27 $\pm$ 0.134 & 0.658 $\pm$ 0.071 & 0.372 $\pm$ 0.117 \\
6 & 0.742 $\pm$ 0.0736 & 0.926 $\pm$ 0.091 & 0.822 $\pm$ 0.055 \\
7 & 1 $\pm$ 0 & 0.318 $\pm$ 0.097 & 0.476 $\pm$ 0.111 \\
8 & 0.96 $\pm$ 0.05 & 0.94 $\pm$ 0.054 & \textbf{0.946 $\pm$ 0.034} \\
9 & 0.68 $\pm$ 0.149 & 0.384 $\pm$ 0.072 & 0.486 $\pm$ 0.091 \\
\hline
\end{tabular}
\label{table_7}
\end{table}

\begin{table}[!htbp]
\centering
\caption{Emerging sensor fault detection with GTNP.}
\begin{tabular}{ccccccc}
\hline
Fault type & $pre$ & $rec$ & $f1$ \\
\hline
0 & 0.814 $\pm$ 0.084 & 0.934 $\pm$ 0.041 & \textbf{0.868 $\pm$ 0.05} \\
1 & 0.984 $\pm$ 0.032 & 0.852 $\pm$ 0.179 & \textbf{0.902 $\pm$ 0.119} \\
2 & 0.956 $\pm$ 0.088 & 0.356 $\pm$ 0.18 & \textbf{0.486 $\pm$ 0.145} \\
3 & 0.664 $\pm$ 0.161 & 0.782 $\pm$ 0.097 & \textbf{0.712 $\pm$ 0.126} \\
4 & 0.812 $\pm$ 0.12 & 0.784 $\pm$ 0.058 & \textbf{0.79 $\pm$ 0.064} \\
5 & 0.652 $\pm$ 0.164 & 0.746 $\pm$ 0.163 & \textbf{0.694 $\pm$ 0.167} \\
6 & 1 $\pm$ 0.17 & 0.86 $\pm$ 0.17 & \textbf{0.912 $\pm$ 0.166} \\
7 & 0.918 $\pm$ 0.167 & 0.756 $\pm$ 0.167 & \textbf{0.83 $\pm$ 0.173} \\
8 & 0.988 $\pm$ 0.182 & 0.876 $\pm$ 0.175 & 0.924 $\pm$ 0.176 \\
9 & 0.58 $\pm$ 0.15 & 0.75 $\pm$ 0.144 & \textbf{0.648 $\pm$ 0.127} \\
\hline
\end{tabular}
\label{table_8}
\end{table}

~\\
~\\
~\\
~\\
\section{Discussion and conclusions}
In this work, we proposed a scalable and reliable DTL method named GTNP for IFD. The embedding of detection knowledge in the source domain by GCN improves the detection accuracy of GTNP method in the target domain. The reference set is an approach that is regarded as an effective way to diminish the demand for samples needed in the target domain. The global and local multi-scale modeling strategy promotes GTNP to better learn the sample features in the source domain and target domain, and improve the detection performance when samples in the target domain are limited. In addition, the multi-scale modeling strategy is an effective method to achieve uncertainty analysis at the model level and sample level. Promisingly, global uncertainty indicates the stability and robustness of the overall model which provides a measure to design the model coping with different conditions or environments. Local uncertainty provides extra uncertainty analysis compared to the traditional methods which facilitates the model to give more accurate outputs for specific samples.

We conduct comprehensive experiments across 3 IFD tasks: the detection of rolling bearing faults under varying speed conditions, the detection of sensor faults between different aircraft, and the detection of emerging fault exclusive to the target domain. The results indicate that the GTNP leads to performance improvements of 14.75\% and 68.05\% compared to the baseline method on the CWRU dataset and aircraft dataset, respectively. Ablation experiments conducted on the aircraft dataset demonstrate the effectiveness of the GCN-based knowledge embedding and feature-based transfer strategy integrated into GTNP. Furthermore, for tasks including only a limited number of emerging fault in the target domain, GTNP ultimately achieves an average $f1$ values of 87\%. Lastly, the analysis of multi-scale uncertainties in GTNP combines the global and the local perspective, which provides a comprehensive understanding of the model and detection results and improves the reliability of the DTL-based methods for real IFD tasks.

The proposed GTNP serves as a valuable benchmark for scalable and reliable DTL methods. It is not only applicable to IFD but also holds promise for a wide range of scenarios involving uncertainty analysis, including autonomous driving, medical detection, etc. Furthermore, the incorporation of the reference set and the multi-scale modeling method endow GTNP with strong performance, particularly in scenarios where target domain samples are limited. Employing the GTNP as the foundation, the modeling approach based on a distribution perspective enhances the neural network's capacity to provide richer analytical information, including interpretability and uncertainty analysis, which contributes to the advancement of trustworthy intelligent detection systems.

In our further work, we will study how to provide more prior knowledge to GTNP (which can be the results of pre-training from other models, inherent human experience, etc.) to improve the detection performance. In addition, we will explore the application of GTNP in other real-world scenarios that require DTL and uncertainty analysis. We also plan to study how to deploy GTNP on low-resource devices to learn trustworthy intelligent transfer models to better serve real-world tasks.

However, our work also has the following limitations. Firstly, the variable distributions constructed are all assumed to be Gaussian distributions, which may not be applicable to special problems. Therefore, more refined data feature engineering and analysis are needed to study what distribution variables obey to better model the target tasks. Secondly, we assume that the labels of the data are accurate. If the labels are ‘‘contaminated’’, the model may be guided by wrong information. Finally, the introduction of the GCN will slightly increase the computational cost, which may not be suitable for edge and mobile devices. Therefore, in practical applications, the model can be further lightweight, or computational acceleration methods such as low-bit quantization can be combined with GTNP to reduce computational costs.

\section*{Acknowledgments}
This work was sponsored by Shanghai Sailing Program (20YF1402500) and Natural Science Foundation of Shanghai (22ZR1404500).

\section*{Data availability}
The CWRU datasets that support the findings of this study are all available ones, and the use of them in this work adheres to the licenses of these datasets. The CWRU dataset is available at \href{https://zenodo.org/records/10405596}{https://zenodo.org/records/10129942}. However, We are not authorized to publicly release the whole dataset used during the current study concerning aircraft sensor faults. Nonetheless, the processed example samples are available on \href{https://zenodo.org/records/10405596}{https://zenodo.org/records/10129942}. Source data are provided with this paper.

\bibliographystyle{elsarticle-num}

\bibliography{bibsample}

\end{document}